%% file: neurips_root.tex
\documentclass{article}

\usepackage[nonatbib, preprint]{neurips_2020}

\usepackage[utf8]{inputenc} 
\usepackage[T1]{fontenc}    
\usepackage{hyperref}       
\usepackage{url}            
\usepackage{booktabs}       
\usepackage{amsfonts}       
\usepackage{nicefrac}       
\usepackage{microtype}      
\usepackage{xcolor}
\usepackage[numbers]{natbib}
\usepackage[pdftex]{graphicx}

\usepackage{amsfonts}
\usepackage{amsmath}
\usepackage{algorithm}
\usepackage{algorithmic}

\usepackage{xcolor}
\usepackage{amssymb}

\newtheorem{lemma}{\bf Lemma}
\newtheorem{assumption}{\bf Assumption}
\newtheorem{theorem}{\bf Theorem}

\newcommand{\E}{\mathbb{E}}

\title{Zeroth-order Deterministic Policy Gradient}

\author{%
	Harshat Kumar\\
	Dept. of Electrical and Systems Eng.\\
	University of Pennsylvania\\
	Philadelphia, PA 19104 \\
	\texttt{harshat@seas.upenn.edu} \\
	 \And
	 Dionysios S. Kalogerias \\
	Dept. of Electrical and Systems Eng.\\
	University of Pennsylvania\\
	Philadelphia, PA 19104 \\
	\texttt{dionysis@seas.upenn.edu} \\
	\AND
	George J.~Pappas \\
	Dept. of Electrical and Systems Eng.\\
	University of Pennsylvania\\
	Philadelphia, PA 19104 \\
	\texttt{pappasg@seas.upenn.edu} \\
	\And
	Alejandro Ribeiro \\
	Dept. of Electrical and Systems Eng.\\
	University of Pennsylvania\\
	Philadelphia, PA 19104 \\
	\texttt{aribeiro@seas.upenn.edu} \\
}

\begin{document}
	
	\maketitle
	
	\begin{abstract}
Deterministic Policy Gradient (DPG) removes a level of randomness from standard randomized-action Policy Gradient (PG), and demonstrates substantial empirical success for tackling complex dynamic problems involving Markov decision processes. At the same time, though, DPG loses its ability to learn in a model-free (i.e., actor-only) fashion, frequently necessitating the use of critics in order to obtain consistent estimates of the associated policy-reward gradient.
In this work, we introduce Zeroth-order Deterministic Policy Gradient (ZDPG), which approximates policy-reward gradients via two-point stochastic evaluations of the $Q$-function, constructed by properly designed low-dimensional action-space perturbations. Exploiting the idea of random horizon rollouts for obtaining unbiased estimates of the $Q$-function, ZDPG lifts the dependence on critics and restores true model-free policy learning, while enjoying built-in and provable algorithmic stability.
Additionally, we present new finite sample complexity bounds for ZDPG, which improve upon existing results by up to two orders of magnitude. Our findings are supported by several numerical experiments, which showcase the effectiveness of ZDPG in a practical setting, and its advantages over both PG and Baseline PG.
%
\end{abstract}

	\input{intro1}
\input{section1}

	\input{convergence3}

    \input{numericals}
	\bibliography{references}
	\bibliographystyle{plain}
	
	\appendix
	\input{appendix1}

\end{document}

%% file: intro1.tex
\section{Introduction}

Reinforcement Learning (RL) has proven itself a well-known and effective approach for tackling dynamic optimization problems, and is ubiquitous in many application areas \cite{arulkumaran2017deep}, such as robotic manipulation \citep{kober2013reinforcement}, supply chain management \cite{giannoccaro2002inventory}, games \cite{silver2016mastering, tesauro1995temporal}, and wireless communications \cite{ye2019deep,eisen2019learning}, to name a few.
A general approach for solving RL problems is \textit{policy search}, which relies on directly learning a policy to maximize rewards through interaction with an unknown environment. Hypothesizing that actions are chosen according to a parameterized distribution (a.k.a. a randomized policy), \emph{Policy Gradient (PG)} is a type of policy search algorithm that repeatedly updates its policy 
through a specially-derived stochastic gradient \cite{sutton2000policy}, enabling solutions in complex problems with continuous state-action spaces. Backing its empirical success are a range of theoretical guarantees including guaranteed policy improvement with smoothness assumptions \cite{pirotta2015policy}, sample complexity bounds \cite{zhang2019global}, and variance reduction \cite{papini2018stochastic}.

\textit{Deterministic Policy Gradient (DPG)} considers RL problems by naturally assuming that the policy is a deterministic mapping from states to actions \cite{silver2014deterministic}. This formulation removes policy randomization, otherwise integral to classical PG, and permits the complete characterization of the gradient of the composed policy-reward function with respect to the parameters of the policy. DPG has been shown to outperform its stochastic counterparts empirically on a wide variety of RL problems \cite{lillicrap2015continuous,gu2017deep}. To evaluate the gradient, however, conventional DPG algorithms rely on critics \cite{konda2003onactor}, which use dynamic programming and function approximation to estimate the state-action value, or $Q$-function. Also, guarantees for convergence of $Q$-learning methods exist only for linear functions \cite{bhandari2018finite, dalal2018finite}, which work under the assumption that a rich feature representation of the state is available. Finding such a feature representation is known to itself be a nontrivial problem \cite{yang2019reinforcement}. Furthermore, the bias on the gradient is difficult to characterize \cite{kumar2019sample}, resulting in suboptimal convergence rates \cite{vemula2019contrasting}.

 



In this work, we introduce \textit{Zeroth-order Deterministic Policy Gradient (ZDPG)}, which advocates the use of (zeroth-order) noisy estimates of the $Q$-function for approximating the corresponding (deterministic) policy-reward gradient, in the standard infinite horizon MDP formulation. 
In particular, by exploiting the recently introduced idea of \textit{random horizon rollouts} for obtaining unbiased stochastic estimates of the $Q$-function \cite{paternain2018stochastic, zhang2019global}, we show that it is possible to efficiently construct consistent two-point approximations (single or in batches) of the associated policy-reward gradient \cite{duchi2015optimal,nesterov2017random}; this is achieved by evaluating pairs of the $Q$-function at low-dimensional random action-space perturbations, involving \textit{initial actions only}.
Such a construction is rather important from a computational perspective, for two main reasons. First, for most popular choices of policy parameterizations (e.g., Deep Neural Networks (DNNs)), the dimension of the corresponding parameter space is typically much larger than the dimension of the action space, where our random perturbations are introduced \cite{vemula2019contrasting}. This is a key feature, because it is well-known that the dimension of the perturbation space has drastic negative effects on the performance of zeroth-order methods \cite{duchi2015optimal,nesterov2017random}.
Second, the fact that we (randomly) perturb only the initial action of the underlying dynamic program makes our ZDPG approach readily implementable, and easily applicable to the infinite horizon setting. 
As compared with standard PG and DPG, ZDPG enjoys the following additional operational advantages: 
\begin{itemize}
\item As with DPG, ZDPG eliminates the need for randomized policies. Although we are not the first to achieve this goal \cite{silver2014deterministic}, we would still like to perpetuate the narrative that randomized policies, although most common, are \textit{not} really necessary in PG-based RL.
\item \textit{At the same time}, while ZDPG exploits gradient information of the adopted (and known) policy parameterization, it \textit{restores} true \textit{model-free} policy learning (as in standard PG); this is naturally lost in DPG, where the use of an auxiliary critic is typically invoked \cite{silver2014deterministic}. 
\item ZDPG enjoys built-in and provable quasi-gradient variance stability (albeit at the expense of an additional system rollout), which is very similar in principle to (heuristic) baseline PG techniques \cite{greensmith2004variance}.
\end{itemize}

We also present detailed analysis characterizing the convergence rate of our proposed ZDPG algorithm under common problem regularity conditions. From a technical standpoint, our analysis is different from that of conventional zeroth-order methods for nonconvex optimization -- which follows the general structure of defining a smoothed surrogate objective (reward), analyzing the convergence of the resulting smoothed problem, and then relating those results back to the original problem (see, e.g., \cite{kalogerias2019zeroth,kalogerias2019model}).
\begin{table}
  \caption{Summary of Rate Results (for number of iterations $T$ and accuracy $\epsilon$)}
  \label{sample-table}
  \centering
  \begin{tabular}{c|cc|cc}
    \toprule
    &\multicolumn{2}{c|}{
    Grad-Lipschitz $Q$-function
    }
    &\multicolumn{2}{c}{
    Hessian-Lipschitz $Q$-function
    }  \\
    \midrule
 Number of Rollouts  ($N$) 
 & \hspace{6pt}   $1$  
 & \hspace{6pt} $T$ 
 & \hspace{6pt} $1$  
 & \hspace{6pt} $\sqrt{T}$  \\
 Smoothness Parameter  ($\mu$) 
 & \hspace{6pt} $T^{-1/4}$  
 & \hspace{6pt} $T^{-1/2}$ 
 & \hspace{6pt} $T^{-1/6}$  
 & \hspace{6pt} $T^{-1/4}$  \\
 Stepsize  ($\alpha$) 
 & \hspace{6pt}  $T^{-3/4}$  
 & \hspace{6pt} $T^{-1/2}$ 
 & \hspace{6pt} $T^{-2/3}$  
 & \hspace{6pt} $T^{-1/2}$ \\
 \midrule
Convergence Rate 
& \hspace{6pt} $\mathcal{O}(\epsilon^{-4})$ 
& \hspace{6pt} $\mathcal{O}(\epsilon^{-2})$ 
& \hspace{6pt} $\mathcal{O}(\epsilon^{-3})$ & \hspace{6pt} $\mathcal{O}(\epsilon^{-2})$   \\
    \bottomrule
  \end{tabular}
  \vspace{-8pt}
\end{table}
More specifically, our convergence analysis is based on 
the explicit construction of a smoothed policy-reward gradient surrogate (not necessarily corresponding to the gradient of some smoothed objective surrogate), which admits a zeroth-order representation matching exactly our rollout-pair-based two-point policy-reward gradient approximations. A key feature of the proposed gradient surrogate is that it constitutes a uniform approximation to the true (deterministic) policy-reward gradient of Silver \textit{et al.} \cite{silver2014deterministic}. This fact is then exploited to analyze the resulting stochastic quasi-gradient algorithm (ZDPG) as a method for solving the original DPG problem, directly.

Our rate results are summarized in Table \ref{sample-table}. Specifically, we show that under the assumption of a Lipschitz and smooth $Q$-function, ZDGP with a single rollout pair achieves a rate of order $\mathcal{O}(\epsilon^{-4})$ (given accuracy $\epsilon>0$), whereas if the $Q$-function has also Lipschitz Hessian, ZDPG achieves a rate of order $\mathcal{O}(\epsilon^{-3})$. If, further, multiple rollout pairs are available (mini-batch equivalent), then the aforementioned rates both improve to an order of $\mathcal{O}(\epsilon^{-2})$, in which case the number of rollout pairs are precisely $T$ and $\sqrt{T}$, respectively, where $T$ is the total number of ZDPG iterations. To the best of our knowledge, our results are the first of their kind pertinent to the standard infinite horizon setting, also outperforming previous results reported for the finite horizon setting \cite{vemula2019contrasting} by up to two orders of magnitude (depending, of course, on problem conditioning). 

The performance of ZDPG is empirically evaluated on a point agent navigation problem at the presence of environment obstacles, where there is no knowledge of the system dynamics. Our numerical results demonstrate that ZDPG substantially outperforms both standard PG and PG with a baseline, under both noiseless and noisy system dynamics, and in terms of both mean performance and its variance. In particular, our simulations elucidate the fact that ZDPG enjoys clear operational advantages over PG with a baseline, which may be considered as the analogous two-point update for the stochastic policy setting. 

%% file: section1.tex
\section{Deterministic Policy Gradient} \label{problem_formulation}

We consider a Reinforcement Learning (RL) problem where an agent moves through a continuous and compact state space $\mathcal{S}\subset \mathbb{R}^q$, and takes actions in a finite dimensional action space $\mathcal{A}=\mathbb{R}^p$.
The agent's task is to accumulate as much reward as possible in the long term, where the reward is revealed to the agent by the environment at each step. This generic problem may be abstracted by a Markov decision process (MDP) as a tuple $(\mathcal{S}, \mathcal{A}, P, R, \gamma)$, where $R:\mathcal{S}\times \mathbb{R}^p \to \mathbb{R}$ is the reward function, and $P_{s\to s'}^{a}:= p(s'\vert (s,a) \in \mathcal{S} \times \mathcal{A})$ determines the probability of moving to state $s'$ starting from state $s$ and action $a$, satisfying the Markov property $p(s_{t+1}\vert (s_u,a_u) \in \mathcal{S} \times \mathcal{A}, \forall u \leq t) = p(s_{t+1}\vert (s_t,a_t) \in \mathcal{S} \times \mathcal{A})$. 
The value $\gamma \in (0,1)$ is the discount factor which determines how much future rewards matter to the behavior of the agent. Lastly, the \textit{deterministic} policy $\pi: \mathcal{S} \to \mathcal{A}$ defines a mapping from a state to an action. Then, the problem of interest is to choose a policy that maximizes the expected discounted sum of rewards over all starting states; this means finding a policy that maximizes the expected value function $V: \mathcal{S} \to \mathbb{R}$ starting from state $s$ or, formally, 
\begin{equation} \label{eqn:problem_main}
\max_{\pi} \E_{s\sim\rho^0} \left[V(s) \hspace{-2pt}:=\hspace{-1pt} \E\left[\sum_{t = 0}^\infty \gamma^t R(s_t,a_t \hspace{-1pt}=\hspace{-1pt} \pi(s_t)) \Bigg| s \right] \right],
\end{equation}
where $\rho^0$ is the initial state distribution. By conditioning the value function on an action, we can define the state-action function $Q: \mathcal{S} \times \mathcal{A} \to \mathbb{R}$ by
\begin{equation} \label{eqn:Q_def}
Q^\pi(s,a) \hspace{-2pt}:=\hspace{-2pt}  R(s,a) \hspace{-1pt}+\hspace{-1pt} \gamma \E_{s'\sim p(\cdot|s,a)}
\hspace{-1.5pt}
\left[ V(s')\right] 
\hspace{-2pt}=\hspace{-1.5pt} \mathbb{E}
\hspace{-2pt}
\left[ \hspace{-1pt} R(s_0, a_0) \hspace{-1.5pt}+\hspace{-1.5pt} \sum_{t=1}^\infty \hspace{-1pt} \gamma^t R(s_t,\pi(s_t)) \bigg|  s_0 \hspace{-1pt}=\hspace{-1pt} s, a_0 \hspace{-1pt}=\hspace{-1pt}  a\right]\hspace{-2pt}.\hspace{-1pt}
\end{equation}
The $Q$-function determines the quality of taking an action $a$ at state $s$ and then following policy $\pi$ for the remainder of time. By selecting the first action according to the policy, the $Q$-function is equivalent to the value function, and hence the problem \eqref{eqn:problem_main} is equivalently written as
\begin{equation} \label{eqn:problem_Q}
\max_\pi \E_{s\sim \rho^0} \left[Q(s,\pi(s))\right].
\end{equation} 
Solving \eqref{eqn:problem_Q} requires searching over a function space, which is generally an intractable problem. To circumvent this complexity, the policy is parameterized by some $\theta \in \Theta \subset \mathbb{R}^d$ so that the search is over a Euclidean space of finite dimension. The problem of interest consequently becomes
\begin{equation} \label{orig_problem}
\max_{\theta }  \mathbb{E}_{s\sim \rho^0}\left[ Q^{\pi_\theta}(s,\pi_{\theta}(s))\right] =: J(\theta).
\end{equation}
%
In their seminal work \cite{silver2014deterministic}, Silver \emph{et al.} proved that, under mild problem regularity conditions, the gradient of \eqref{orig_problem} can be expressed by
\begin{equation} \label{eqn:grad_obj}
\nabla  J(\theta) = \mathbb{E}_{s\sim \rho^\pi}\left[\nabla_{\theta}\pi_\theta(s) \nabla_a  Q^{\pi_\theta}(s,a)\vert_{a = \pi_\theta(s)}\right],
\end{equation}
where $\rho^\pi$ is the so-called (improper) \textit{discounted state distribution} under policy $\pi$, defined as $\rho^\pi(s') =  \int_{\mathcal{S}} \sum_{t = 0}^\infty \gamma^t p_0(s)p(s\to s', t, \theta) ds$, and $p(s\to s', t,\theta)$ denotes the density at state $s'$ after starting form state $s$ and transitioning with $t$ time steps given policy parameterized by $\theta$. Next, we formally define some standard assumptions on the policy gradient, reward function, and $Q$-function, as follows.
\begin{assumption}\label{asm:bounded_gradient} \textbf{ (Bounded Policy Gradient)}
	The policy gradient $\nabla_\theta \pi_\theta(s)$ is uniformly bounded   on $\Theta\times\mathcal{S}$, i.e., there is $B_\Theta<\infty$, such that $\sup_{(\theta,s)\in\Theta\times\mathcal{S}}\|\nabla_\theta \pi_\theta(s) \| \leq B_\Theta$.
\end{assumption}

\begin{assumption} \label{asm:bounded_reward}\textbf{(Bounded Rewards)}
	The reward $R$ is uniformly bounded on $\mathcal{S}\times \mathcal{A}$, i.e., there is $U_R<\infty$, such that $\sup_{(s,a)\in \mathcal{S}\times \mathcal{A}}|R(s,a)|\le U_R$.
	%
\end{assumption} 
By Assumption \ref{asm:bounded_reward}, it also follows that the $Q$-function is also uniformly bounded as
\begin{equation*} 
|Q^{\pi_\theta}(s,a)| \leq \sum_{t=0}^\infty \gamma^t \cdot U_R = \frac{1}{1-\gamma} U_R =: \mathcal{Q}.
\end{equation*}
Both Assumptions \ref{asm:bounded_gradient} and \ref{asm:bounded_reward} are standard for deriving performance guarantees in the policy search literature \cite{zhang2019global,kumar2019sample, bhandari2018finite}, and are easily satisfied by common choices of the deterministic policy and reward functions. Additionally, we require sufficient smoothness of the $Q$-function, as follows. 
\begin{assumption} \label{asm:smooth} \textbf{(Lipschitzness of the $Q$-function)} The $Q$-function is action-Lipschitz uniformly on $\Theta \times \mathcal{S}$. That is, there is a number $L <\infty$, such that, for every $a_1, a_2 \in \mathcal{A}$,
	\begin{equation*}
	\sup_{(\theta, s)\in \Theta \times\mathcal{S}}| Q^{\pi_{\theta}}(s,a_1) -  Q^{\pi_{\theta}}(s,a_2)| \leq L \|a_1 - a_2\|.
	\end{equation*}
	%
\end{assumption}
\begin{assumption} \label{asm:nesterov_assumptions} \textbf{(Smoothness of the $Q$-function)} The $Q$-function is action-smooth uniformly on $\Theta \times \mathcal{S}$.  That is, there is a number $G <\infty$, such that, for every $a_1, a_2 \in \mathcal{A}$,
	\begin{equation*}
	\sup_{(\theta, s)\in \Theta \times\mathcal{S}}\|\nabla_a Q^{\pi_{\theta}}(s,a_1) - \nabla_a Q^{\pi_{\theta}}(s,a_2)\| \leq G \|a_1 - a_2\|.
	\end{equation*}
	%
	%
	%
\end{assumption}
Intuitively, Assumptions \ref{asm:smooth} and \ref{asm:nesterov_assumptions} means that similar actions from the same state should have similar values. This principle has informed Actor-Critic algorithms in practice as they are known to avoid overfitting to narrow peaks in the value estimate \cite{fujimoto2018addressing}. Furthermore, existing convergence analysis for zeroth-order action space exploration with finite horizon use these assumptions as well \cite{vemula2019contrasting}. Lastly, in some of our results we assume that the problem is well-conditioned with Lipschitz Hessians.
\begin{assumption} \label{asm:nesterov_assumptions_H} \textbf{(Smoothness of $\nabla Q$)} The gradient field $\nabla Q$ is action-smooth uniformly on $\Theta \times \mathcal{S}$.  That is, there is a number $H <\infty$, such that, for every $a_1, a_2 \in \mathcal{A}$,
	\begin{equation*}
	\sup_{(\theta, s)\in \Theta \times\mathcal{S}}\|\nabla_a^2 Q^{\pi_{\theta}}(s,a_1) - \nabla_a^2 Q^{\pi_{\theta}}(s,a_2)\| \leq H \|a_1 - a_2\|.
	\end{equation*}
	%
	%
	%
\end{assumption}
Assumption \ref{asm:nesterov_assumptions_H}, though not entirely standard, holds for a variety of problems with well behaved reward functions and policy representations, such as the Linear Quadratic Regulator (LQR) \cite{fazel2018global, malik2020derivative, mohammadi2019convergence} and its many (and potentially nonlinear) variations. A reward function may be designed to achieve such an assumption, as well \cite{ng1999policy}. The trade-off between our various smoothness assumptions and the convergence rates achieved in this work is demonstrated in Table \ref{sample-table}. 

In order to compute a stochastic approximation of \eqref{eqn:grad_obj}, we need to both sample from the discounted state distribution $\rho^\pi(s)$ and evaluate the gradient expression. Because the policy is deterministic, $\nabla_\theta\pi_\theta(s)$ is the Jacobian of the policy $\pi_\theta$ evaluated at $s$ which can be explicitly calculated. Instead of using a critic network to estimate the $\nabla_a Q^{\pi_\theta}(s,a)$ \cite{silver2014deterministic}, which requires many state-action-reward samples before converging to something meaningful, we propose the use of two point zeroth-order action space perturbed $Q$-function estimates. In the following section, we review how to use zeroth-order information to evaluate stochastic gradient estimates.  

\section{Zeroth-order Gradient Estimates and ZDPG}
%
First, we introduce a smoothed approximation to the $Q$-function by exploiting low-dimensional perturbations with respect to the action $a$.
%
%
%
Let $\mu > 0$ be \textit{the smoothing parameter}, and let $\mathbf{u}\sim \mathcal{N}(0, I_p)$. Then, the $\mu$-smoothed $Q$-function is defined as 
\begin{equation*}
Q_\mu^{\pi_\theta}(s,a) := \mathbb{E}_\mathbf{u}\left[Q^{\pi_\theta}(s,a + \mu \mathbf{u})\right].
\end{equation*}
By introducing smoothing, we are able to sample an unbiased stochastic estimate of $\nabla_a Q_\mu^{\pi_\theta}(s,a)$ via a two-point evaluation of the original $Q$-function.  Namely, we recall the following key property (in terms of the $Q$-function), true in general for every globally Lipschitz function as originally presented in \cite{nesterov2017random}, then extended to a wider class of functions in \cite{kalogerias2019zeroth}.
\begin{lemma} \emph{\cite[Lemma 2]{kalogerias2019zeroth}} \label{lem:unbiased_smoothed_estimate}
	For every $\mu > 0$, the $\mu$-smoothed $Q$-function surrogate $Q_\mu^{\pi_\theta}$ is differentiable, and its gradient admits the representations 
	\begin{align}
	\nabla_a Q_\mu^{\pi_\theta}(s,a) &\equiv \E_{\mathbf{u} \sim \mathcal{N}(0, I_p)} \left[ \frac{Q^{\pi_{\theta}}(s,a + \mu \mathbf{u} ) - Q^{\pi_{\theta}}(s,a)}{\mu}\mathbf{u} \right] \nonumber \\
	&\equiv \E_{\mathbf{u} \sim \mathcal{N}(0, I_p)} \left[ \frac{Q^{\pi_{\theta}}(s,a + \mu \mathbf{u} ) - Q^{\pi_{\theta}}(s,a - \mu \mathbf{u})}{2\mu}\mathbf{u} \right]. \label{equ:symm_diff}
	\end{align}
\end{lemma}
Driven by Lemma \ref{lem:unbiased_smoothed_estimate}, we propose the following estimate of the objective gradient defined in \eqref{eqn:grad_obj} where, instead of the true gradient $\nabla_a Q^{\pi_
\theta}(s,a)$ we use the smoothed gradient $\nabla_a Q^{\pi_\theta}_\mu (s,a)$ to obtain the quasi-gradient
\begin{equation} \label{eqn:grad_smooth_obj}
    \hat \nabla J(\theta) = \E_{s\sim \rho^{\pi}} \left[\nabla_\theta \pi_\theta(s)\nabla_a Q^{\pi_\theta}_\mu (s,a) \vert_{a = \pi_\theta(s)}\right].
\end{equation}
Lemma \ref{lem:unbiased_smoothed_estimate} provides a procedure to obtain an unbiased estimate of $\nabla_a Q_\mu^{\pi_{\theta}}(s,a)$, which is to sample $\mathbf{u}\sim \mathcal{N}(0, I_p)$ and then evaluate the $Q$-function with \textit{initial} actions $a$ and $a + \mu \mathbf{u}$. By selecting the smoothing parameter $\mu$ sufficiently small, the gradient estimate becomes close to the deterministic policy gradient (see Section \ref{section_convergence}, Theorem \ref{thm:nesterov_result1}). Obtaining two samples of the $Q$-function at a specific state may not be practical on a real system, but can be made feasible through the use of simulators which allow resetting and reproduction of the same stochastic environment \cite{schulman2015trust}. 
The resulting stochastic search algorithm using the two point zeroth-order gradient estimate is presented in pseudocode in Algorithm \ref{alg:ZOPG}, where $\alpha$ is the policy parameter stepsize. We employ Monte-Carlo variance reduction by running the system $N\ge1$ times ($N=1$ implies no reduction).


\begin{algorithm}[t]
	\caption{Zeroth-order Deterministic Policy Gradient}
	\begin{algorithmic}	\label{alg:ZOPG}
		\REQUIRE $\theta_0$, $  \gamma, \mu, \alpha, N$
		\FOR{$t = 0, 1, \dots$}
		\STATE Sample $T_Q \sim \textrm{Geom}(1-\gamma)$
		\STATE Sample $s_t\sim (1-\gamma)\rho^{\pi_{\theta_t}}$ [Algorithm \ref{alg:sample_rho} with $(T,\theta) = (T_Q,\theta_t)$]
		\STATE Sample $\mathbf{u} \sim \mathcal{N}(0,I_p)$
		\STATE Initialize Estimates $\hat Q^+_{t} \leftarrow 0$, $\hat Q^-_{t} \leftarrow 0$
		\FOR{$n = 0, 1, \dots, N-1$}
		\STATE $\hat Q^+_{t,n}$ $\leftarrow$ Algorithm \ref{alg:estimateQ} with $(T, s_0, a_0,\theta) = (T_Q, s_t,  \pi_{\theta_t}(s_t) + \mu \mathbf{u},\theta_t)$ 
		\STATE $\hat Q^-_{t,n}$ $\leftarrow$ Algorithm \ref{alg:estimateQ} with $(T, s_0, a_0,\theta) = (T_Q, s_t,  \pi_{\theta_t}(s_t),\theta_t)$
		\STATE $\hat Q^+_{t} \leftarrow  \hat Q^+_{t} + \frac{1}{N} \hat Q^-_{t,n}$  
		\STATE $\hat Q^-_{t} \leftarrow  \hat Q^-_{t} + \frac{1}{N} \hat Q^-_{t,n}$  
		\ENDFOR
		\STATE Evaluate $\Psi_t = \nabla_\theta \pi_{\theta_t} (s_t)$
		\STATE $g_t \leftarrow 
		\frac{1}{1-\gamma}
		\Psi_t \left(\frac{\tilde Q^+_t - \tilde{Q}^-_t}{\mu}\right)\mathbf{u}$
		\STATE $\theta_{t+1} \leftarrow \theta_t + \alpha g_t$
		\ENDFOR
	\end{algorithmic}
\end{algorithm}
Algorithm \ref{alg:ZOPG} functions with the assumption that both an unbiased estimate of the $Q$-function evaluated at any state-action pair as well as sampling from the discounted state distribution $\rho^{\pi_\theta}$ are feasible. Here, we describe both procedures which involve sampling a horizon length from a geometric distribution. In \cite[Proposition 2]{paternain2018stochastic}, it was shown that the procedure described by Algorithm \ref{alg:estimateQ} results in unbiased estimates of the $Q$-function defined by \eqref{eqn:Q_def}. As the next result asserts, the aforementioned estimates are of finite variance, as well. Due to lack of space, we defer all subsequent proofs of our theoretical results to the supplementary material.

%
%
%
\begin{lemma} \label{lem:Bounded_Q_var}
	Let Assumption \ref{asm:bounded_reward} be in effect. Then, the estimate $\hat Q$ produced by Algorithm \ref{alg:estimateQ} is of bounded variance, independent of the inputs of Algorithm \ref{alg:estimateQ}.
\end{lemma}
For later reference, let $w^\pm_{t,n}  := \hat Q^\pm_{t,n} - Q^{\pi_{\theta_t}}(\pm)$, where "$\pm$" refers to the respective initial state-action pair. Also introduced in \cite{paternain2018stochastic}, Algorithm \ref{alg:sample_rho} describes a similar procedure to obtain samples $s \sim (1-\gamma)\rho^{\pi_{\theta}}$, where $(1-\gamma)\rho^{\pi_{\theta}}$ is the \textit{proper} (i.e., scaled to sum to unity) version of the discounted state distribution $\rho^{\pi_{\theta}}$.


%% file: convergence3.tex
\section{Convergence Analysis} \label{section_convergence}
Somewhat departing from conventional analysis of zeroth-order methods, we instead bound the bias between our proposed gradient \eqref{eqn:grad_smooth_obj} with the true gradient \eqref{eqn:grad_obj} and characterize the rate using a biased variant of stochastic gradient descent. First, we present the theorem which relates the smoothed objective function gradient $\hat \nabla J(\theta)$ with the true gradient $\nabla J(\theta)$. 
\begin{theorem} \label{thm:nesterov_result1}
	Let Assumptions \ref{asm:bounded_gradient}-\ref{asm:smooth} be in effect. Then, under Assumption \ref{asm:nesterov_assumptions}, it is true that 
	$$\sup_{\theta\in\Theta}\| \nabla J(\theta) - \hat  \nabla J(\theta)\| \leq  B_\Theta \mu \sqrt{p} G/(1-\gamma).$$
	Alternatively, under Assumption \ref{asm:nesterov_assumptions_H}, it is true that
	$$\sup_{\theta \in \Theta}\| \nabla J(\theta) - \hat  \nabla J(\theta)\| \leq 
	B_\Theta \mu^2 (p+4)^2 H/(1-\gamma). $$
\end{theorem}
\begin{figure}[t]
\begin{minipage}{.485\textwidth}
  \begin{algorithm}[H]
	
	\caption{$Q$-function Sampler \cite{paternain2018stochastic}}
	\begin{algorithmic} \label{alg:estimateQ}
		\REQUIRE $\hat{Q} \leftarrow 0$, $T,  s_0, a_0, \theta$
		\FOR{$t = 0, 1, \dots, T - 1$} 
		\STATE Collect Reward $\hat Q \leftarrow \hat Q + R(s_t,a_t)$ 
		\STATE Advance System $s_{t+1} \sim \mathbb{P}(s'\vert s_t,a_t)$
		\STATE Select Action $a_{t+1} \leftarrow  \pi_\theta(s_{t+1})$
		\ENDFOR
		\STATE Collect Reward $\hat{Q} \leftarrow \hat Q + R(s_{T}, a_{T})$
	\end{algorithmic}
\end{algorithm}
\end{minipage}%
\hspace{8.6pt}
\begin{minipage}{.485\textwidth}
  \begin{algorithm}[H]
	\caption{Discounted State Sampler \cite{paternain2018stochastic}}
	\begin{algorithmic}	\label{alg:sample_rho}
		\REQUIRE $T$, $\theta$  $\vphantom{\hat{Q} \leftarrow 0, T,  s_0 = s, a_0 = a, \theta}$
		\STATE Sample $s_0 \sim \rho^{0}$ $\vphantom{\hat{Q}}$
		\FOR{$t = 0, 1, \dots, T - 1$}
		\vspace{0.4pt}
		\STATE Select Action $a_t \leftarrow \pi_\theta(s_t)$
		\STATE Advance System $s_{t+1} \sim \mathbb{P}(s'\vert s_t,a_t)$
		\ENDFOR
		\STATE Return $s_{T}$ $\vphantom{\hat{Q}}$
	\end{algorithmic}
\end{algorithm}
\end{minipage}%
\end{figure}
Assumption \ref{asm:nesterov_assumptions} is key for the former part of Theorem \ref{thm:nesterov_result1}, as the difference between the gradients of the two point $Q$-function with respect to the action $a$ is bounded. The latter part of Theorem \ref{thm:nesterov_result1} is a byproduct of \cite[Lemma 3]{nesterov2017random}. Alternatively, the relation between the estimated gradient with two points and the true gradient could be characterized by \cite[Lemma 3]{nesterov2017random}. Requiring the same assumptions as detailed here except for the bounded objective, this result is not advantageous as it results in a bound of order $\mathcal{O}(p^{3/2})$, contrary to the $\mathcal{O}(p^{1/2})$ bound we provide for this specific case.  

%

By construction of Algorithm \ref{alg:ZOPG}, we have that $g_t$ is an unbiased estimate of the smoothed gradient $\hat \nabla J(\theta)$ defined in \eqref{eqn:grad_smooth_obj}. Next we will show that the second-order moment of $g_t$ is bounded. Doing so will allow us to use analysis similar to \cite{ghadimi2013stochastic} in order to characterize the rate of Algorithm \ref{alg:ZOPG}. 

\begin{lemma} \label{lem:bounded_variance}
Let Assumptions \ref{asm:bounded_gradient}-\ref{asm:smooth} be in effect. Then, the second-order moment of the quasi-gradient $g_t$ in Algorithm \ref{alg:ZOPG} is bounded. In particular, it is true that
 $$\E\big[\hspace{-1pt}\left\|g_t 
 \right\|^2\big] \leq \dfrac{2B_{\Theta}^{2}}{(1-\gamma)^2}\left(\hspace{-1.5pt}(p+4)^2 L^2 + \frac{p \sigma^2}{\mu^2N}\right)=:V,$$
\end{lemma}
where $\infty>\sigma^2\ge\mathbb{E}\{|w_{t,n}^{+}-w_{t,n}^{-}|^{2}|s_t,{\bf u},\theta_t,T_{Q}\}$, deterministic and same for all $t$ and $n$.

We are now ready to state our main theorem, which establishes the convergence rate of Algorithm \ref{alg:ZOPG} with respect to the true gradient of $J(\theta)$.
\begin{theorem} \label{thm:main_1}
Let Assumptions \ref{asm:bounded_gradient}-\ref{asm:smooth} be in effect. Further let the objective $J(\theta)$ be $L_J$-smooth and $G_J$-Lipschitz continuous. Then, under Assumption \ref{asm:nesterov_assumptions}, it is true that 
$$	\frac{1}{T}\sum_{t= 1}^T \E \left[\|\nabla_\theta J(\theta_t)\|^2\right] \leq  \begin{cases} \mathcal{O} \left(p^2T^{-1/4} \right), & \textrm{if~} (N,\alpha,\mu) = (1, T^{-3/4}, T^{-1/4}) \\ \mathcal{O} \left(p^2T^{-1/2} \right), &\textrm{if~} (N,\alpha,\mu) = (T, T^{-1/2}, T^{-1/2})  \end{cases}.$$
Alternatively, under Assumption \ref{asm:nesterov_assumptions_H}, it holds that 
$$	\frac{1}{T}\sum_{t= 1}^T \E \left[\|\nabla_\theta J(\theta_t)\|^2\right] \leq  \begin{cases} \mathcal{O} \left(p^2T^{-1/3} \right), & \textrm{if~} (N,\alpha,\mu) = (1,T^{-2/3}, T^{-1/6}) \\ \mathcal{O} \left(p^2T^{-1/2} \right),  &\textrm{if~} (N,\alpha,\mu) = (T^{1/2},  T^{-1/2},  T^{-1/4})  \end{cases}.$$
\end{theorem}
Theorem \ref{thm:main_1} requires that the objective function $J(\theta)$ is smooth and Lipschitz. This is is easily satisfied under \textit{all} Assumptions \ref{asm:bounded_gradient}-\ref{asm:nesterov_assumptions}, whenever that the adopted policy parameterization is sufficiently well-behaved. A smooth policy serves this purpose, for instance. Further, for simplicity, we have presented Theorem \ref{thm:main_1} for the unconstrained setting, where $\Theta = \mathbb{R}^d$. In practice, though, we may need to constrain the parameters. In this case, Theorem \ref{thm:main_1} continues to hold (\textit{almost}), but with the squared gradient norm being replaced by the merit function $\|\Pi_\Theta\{\theta_t + \alpha \nabla J(\theta_t)\} - \theta_t\|^2 / \alpha^2$; see, e.g., \cite{ghadimi2016mini}.

%% file: numericals.tex
\section{Numerical Results} \label{sec:Numericals}

In this section, we empirically evaluate ZDPG 
and compare its performance to vanilla Policy Gradient (PG) \cite{sutton2000policy} and Policy Gradient with Baseline (PG-B) \cite{greensmith2004variance} with varying levels of system noise and MC variance reduction. First, we formally describe the Markov Decision Process we choose to solve, then we present our results and discuss the key takeaways.

\subsection{Problem Setting: Navigating Around an Obstacle with Unknown Agent Dynamics}
For our empirical evaluation of ZDPG, we consider a two-dimensional continuous state-action navigation problem, where a point agent aims 
to avoid a spherical (for simplicity) obstacle and converge fast to some fixed target while satisfying its motion control dynamics. This goal is achieved by appropriately shaping the associated reward function. 
Let $s^* \in \mathbb{R}^2$ represent the target the point agent wishes to reach. Let $s^c \in \mathbb{R}^2$ and $r\in \mathbb{R}_+$ denote the center and radius of the spherical obstacle the agent wishes to avoid. Then, we define the reward $R: \mathbb{R}^2 \to \mathbb{R}$ as a combination of two artificial potential functions: An attractive potential $\phi^{att}: \mathbb{R}^2\to \mathbb{R}_+$ with a maximum at the target, and a repulsive artificial potential $\phi^{rep}:\mathbb{R}^2 \to \mathbb{R}_+$ with a minimum at the center of the object. Specifically, we set the attractive potential to be the negative Euclidean distance to the goal location $s^*$ squared, that is, $\phi^{att}(s) :=- \|s - s^*\|_2^2$, where $s$ is the location (state) of the agent. Similar to \cite{kalogerias2014mobi}, we define the repulsive potential as $\phi^{rep}(s) := 1 - 1/\beta(s)$, where 
\begin{equation*}
    \beta(s) := \left( 1 - \left(\frac{1 + r^4}{r^4}\right) \cdot \frac{\left(\|s - s^c\|^2 - r^2\right)^2}{1+\left(\|s - s^c\|^2 - r^2\right)^2} \right)^{\frac{1}{2}\left(1 - \textrm{sign}\left(\|s - s^c\| - r\right)\right)}.
\end{equation*}
Completely characterized by the state of the agent, the reward is equal to the sum of these attractive and repulsive potentials, that is, $R(s) = \phi^{att}(s) + \phi^{rep}(s)$. Also, the motion dynamics of the point agent are described by the controlled random walk \begin{equation}\label{state-model}
    s_{t+1} = A s_{t} + \eta \dfrac{\pi_\theta(s_t)}{\Vert \pi_\theta(s_t) \Vert} + n_t,
\end{equation}
where $A\in\mathbb{R}^{2\times2}$ is the motion state transition matrix of the agent, $\{n_t\}$ is a random process modeling state process noise, and $\eta>0$ (together with policy normalization) imposes implicit resource constraints on the agent motion. Here, we assume that, except from the agent control vector (second term on the right-hand side of \eqref{state-model}), the dynamics of the agent are \textit{unknown}, an assumption which conforms with the standard policy gradient setting.
Lastly, we consider a linear, deterministic policy to determine optimal directions that the point agent will follow given its location to its prescribed destination. In particular, we adopt 
a Radial-Basis-Function (RBF) expansion as our (universal) policy parameterization, i.e., 
\begin{equation} \label{equ:det_policy}
    \pi_\theta(s) = \sum_{i = 1}^d \theta_i \kappa(s, \bar s_i),
\end{equation}
%
where $\kappa(s, s') = \textrm{exp}(-\|s-s'\|^2_2/2\sigma_{\kappa}^2)$ defines the RBF kernel, $\theta_i \in \mathbb{R}^2, i = 1,\dots, d$ are the parameter vectors to be learned, $\sigma_{\kappa}^2$ determines the variance of the kernel, and $\bar s_i, i = 1,\dots, d$ are the associated RBF centers. 

For all simulations, we select the target of the agent to be located at $s^* = (-5,-5)$, and the obstacle to be centered at $s^c = (0,0)$ with a radius of $r = 2.5$. We define the potential with a slightly larger radius ($r = 3$) to ensure that the agent does not collide with the boundary of the obstacle. We select the centers of the RBF kernels to be spaced $0.25$ units from each other populating the $[-10,10]\times [-10,10]$ grid. The initial state distribution is uniform on $[-10,10]\times [-10,10]$.

\subsection{Results}
\vspace{-6pt}

\begin{figure*}[!t]
	\centering
	\begin{tabular}{ccc}
		\hspace{3mm}\includegraphics[trim = {5cm, 10cm, 5cm, 10cm}, width=.274\linewidth]{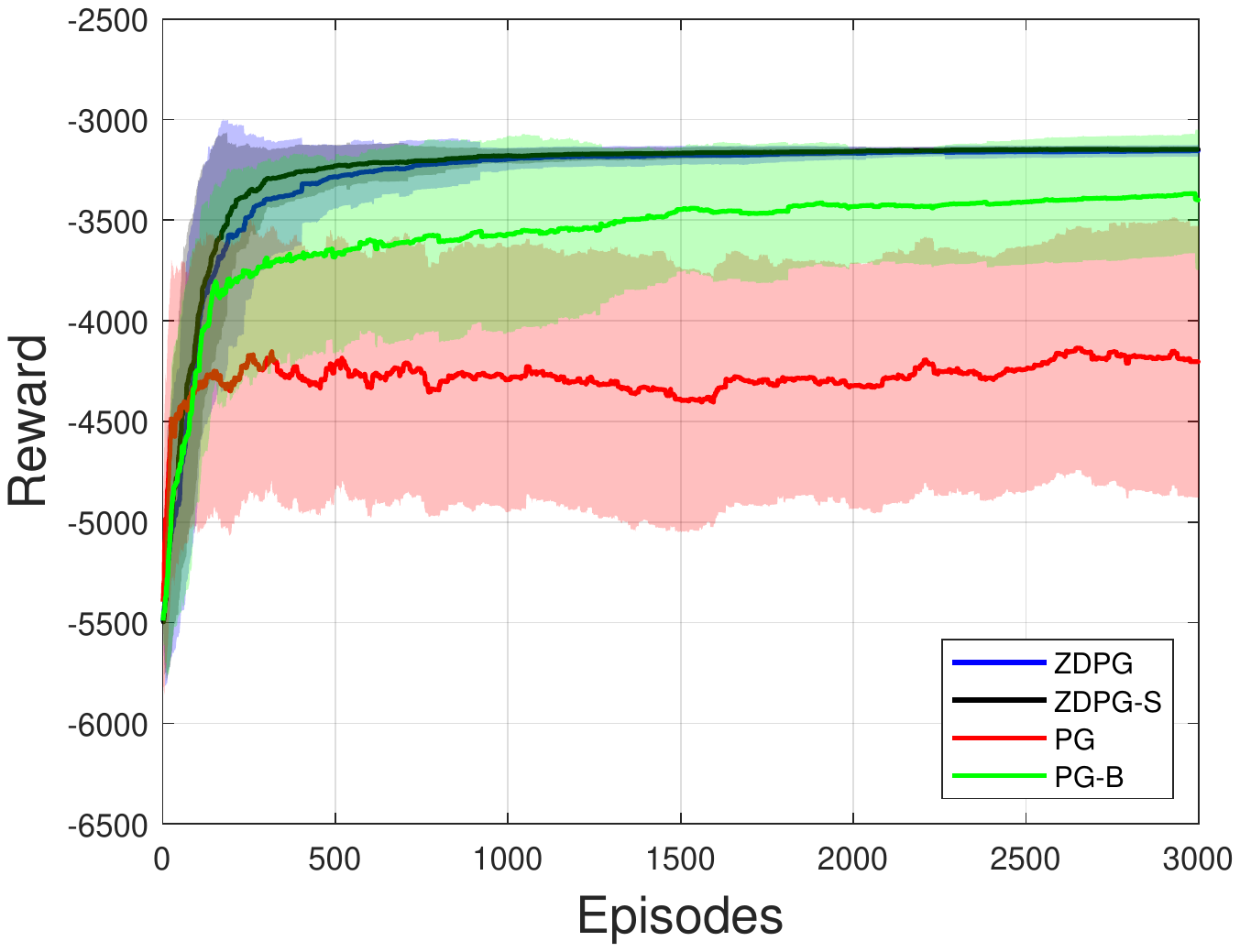}   &\hspace{3.5mm}
		\includegraphics[trim = {5cm, 10cm, 5cm, 10cm},width=.274\linewidth]{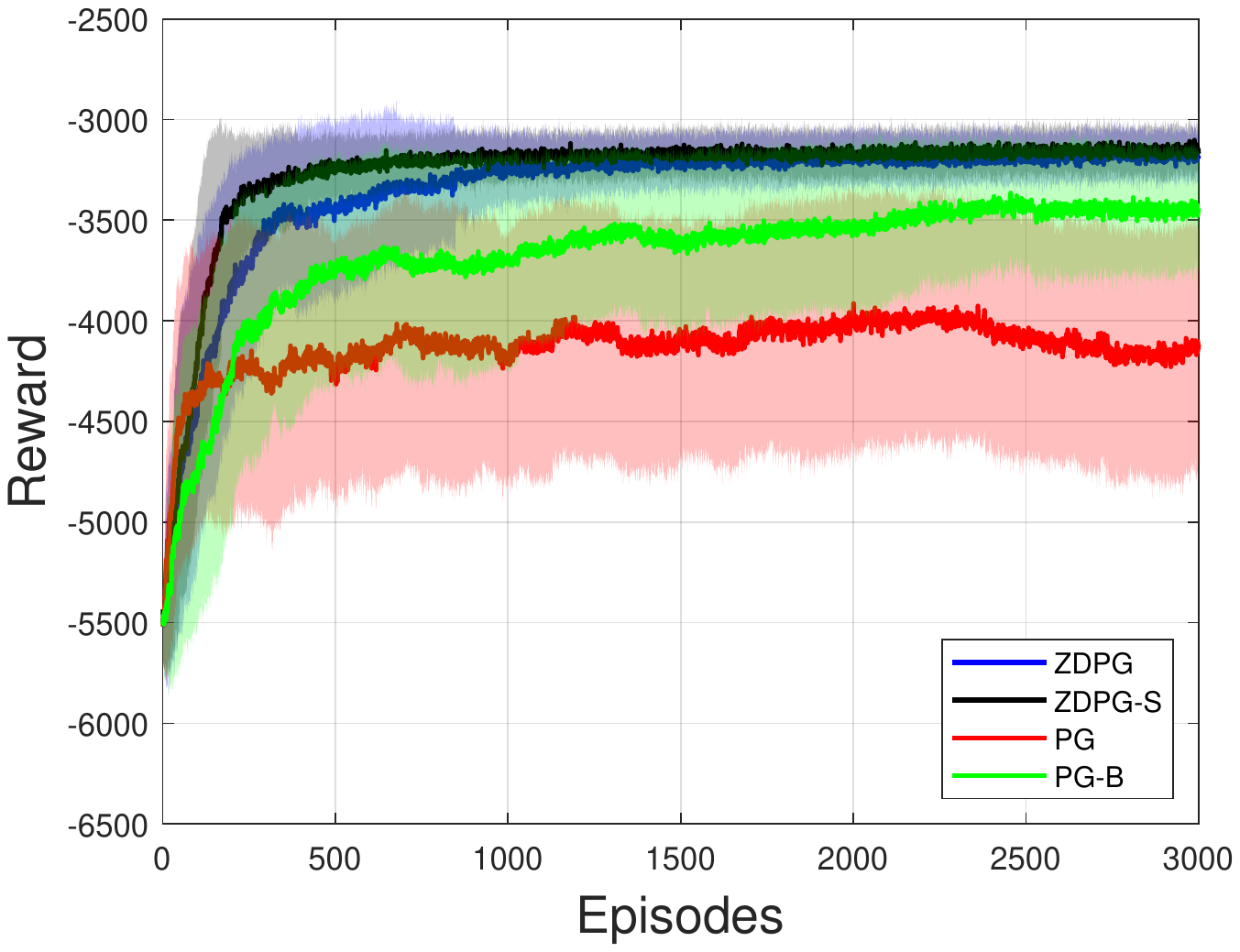} &\hspace{3.5mm}
		\includegraphics[trim = {5cm, 10cm, 5cm, 10cm},width=.274\linewidth]{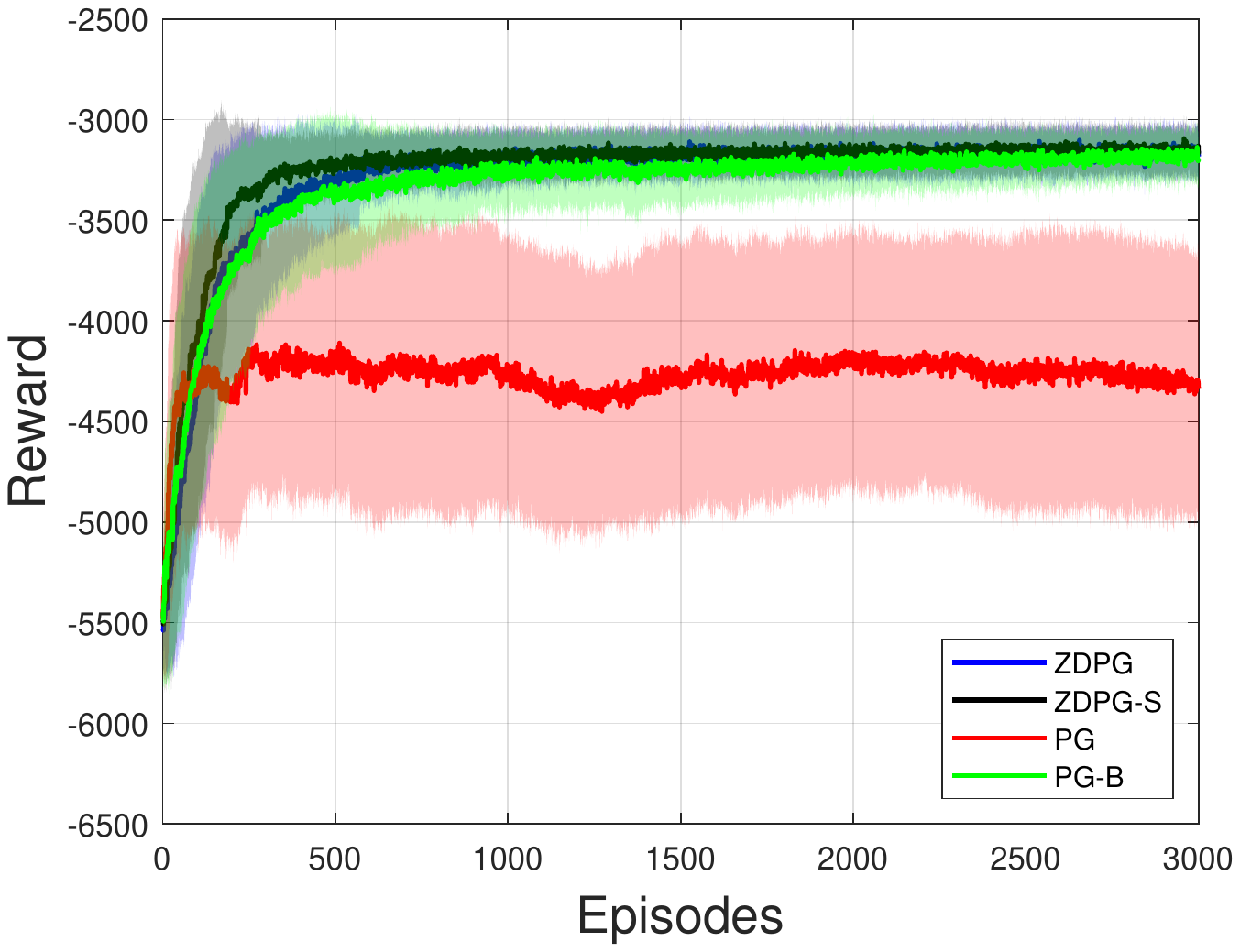} \\ \vspace{.4cm}
		~&~&~\\
		\small (a) & \small (b) & \small (c)
	\end{tabular}
	\caption{Average reward per episode with confidence bounds over 50 trials. Learning rate set to $\alpha = 10^{-7}$ for all simulations. (a) Noiseless case with no MC variance reduction ($N = 1$), ZDPG and ZDPG-S smoothing parameters set to $0.05$ and $0.5$ respectively. Action noise for PG and PG-B set to $\sigma_a^2 = 0.025$. (b) Noisy case ($\omega = 0.01$) with no MC reduction ($N = 1$), ZDPG and ZDPG-S smoothing parameters set to $0.25$ and $0.5$ respectively. Action noise for PG and PG-B set to $\sigma_a^2 = 0.025$. (c) Noisy case (same as (b)) with MC reduction ($N = 10$).}\vspace{2mm}
	\label{fig:LC}
\end{figure*}

The learning curves shown in Figure \ref{fig:LC} illustrate the performance of ZDPG and its symmetric counterpart ZDPG-S (based on \eqref{equ:symm_diff}), compared with PG and PG with a \textit{stochastic} baseline (PG-B), with discount factor $\gamma = 0.8$ in the case with and without system noise ($n_t = \mathbf{0}$ and $n_t \sim \mathcal{N}(\mathbf{0},\omega \cdot I_2)$, respectively). Also, we set $A = I_2$. For PG and PG-B, we let $\tilde \pi$ denote the stochastic policy where the action is selected by $a_t \sim \tilde \pi_\theta(s_t)$. In particular, we let $\tilde \pi$ be a Gaussian policy with mean equal to $\pi_\theta(s)$ [c.f. \eqref{equ:det_policy}] and a standard choice of covariance matrix $\Sigma = \sigma_a^2 \cdot I_2$. 
For PG-B, the baseline is fairly chosen and is analogous as compared to ZDPG in that it scales the update direction $\nabla_\theta \log \tilde \pi_\theta(s)$ by a two-point difference ($\hat Q(s,a) - \tilde{V}(s)$), where $\tilde{V}(s)$ is a single-rollout estimate of the associated value function at state $s$.  
After each policy update, we evaluate the system by generating a rollout of fixed length $T = 20$ from a fixed starting point $s^0 = (5,5)$. As done commonly in practice (as a heuristic), we evaluate the stochastic policies by setting the covariance matrix equal to zero thereby making the evaluation policy deterministic.

\begin{figure*}[!t]
	\centering
	\begin{tabular}{ccc}
		\hspace{.8mm}
		\includegraphics[trim = {7cm, 10cm, 5cm, 10cm}, width=.278\linewidth]{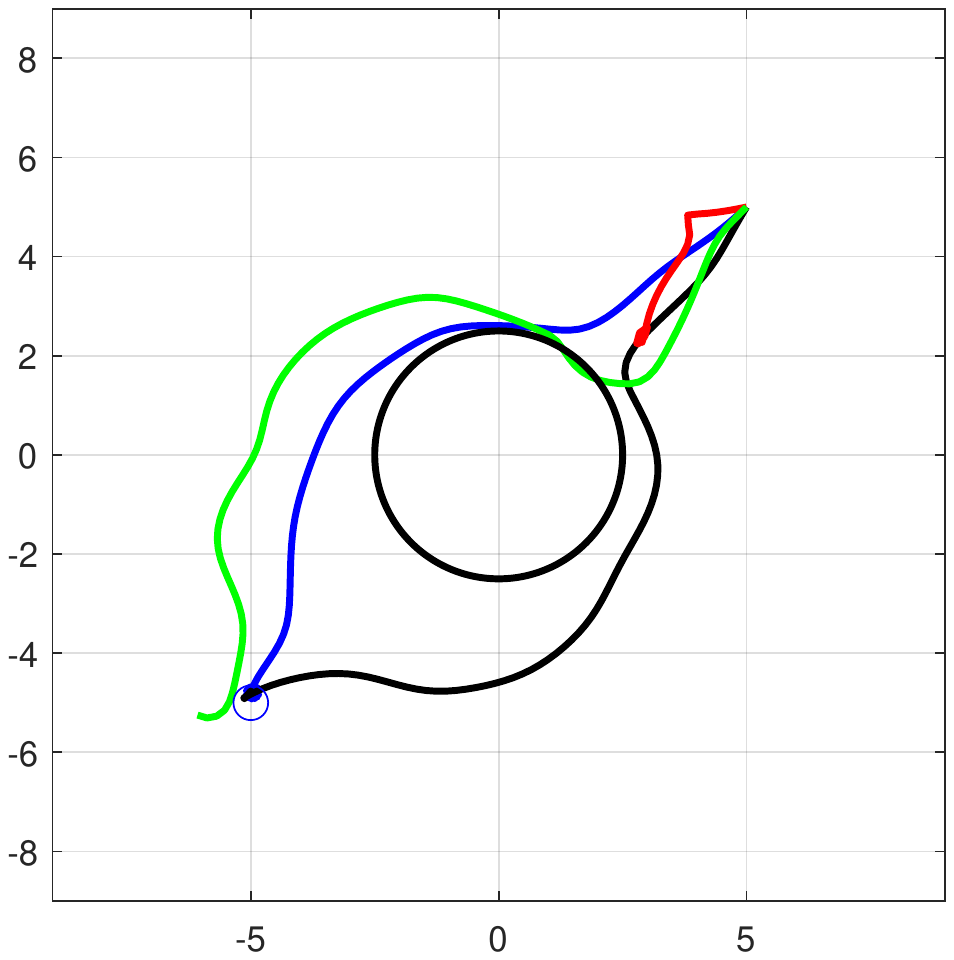}   &\hspace{6mm}
		\includegraphics[trim = {7cm, 10cm, 5cm, 10cm}, width=.278\linewidth]{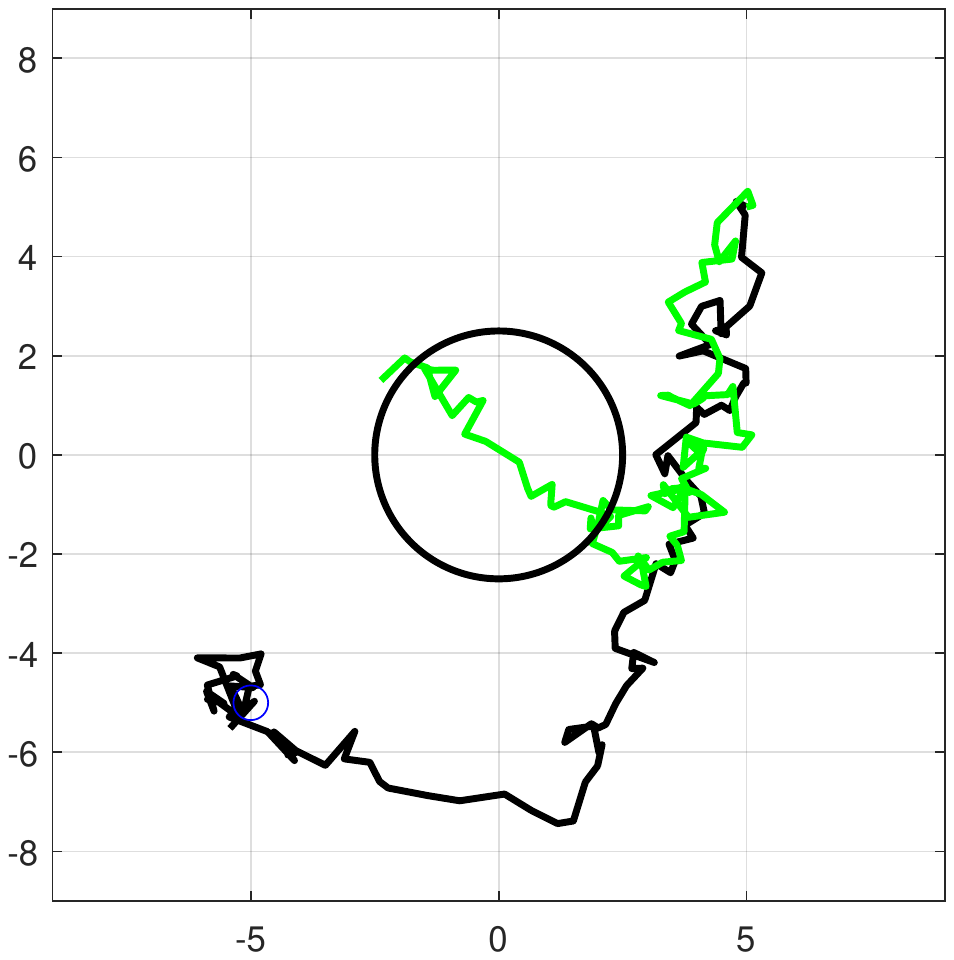} 
		&\hspace{6mm}
		\includegraphics[trim = {7cm, 10cm, 5cm, 10cm}, width=.278\linewidth]{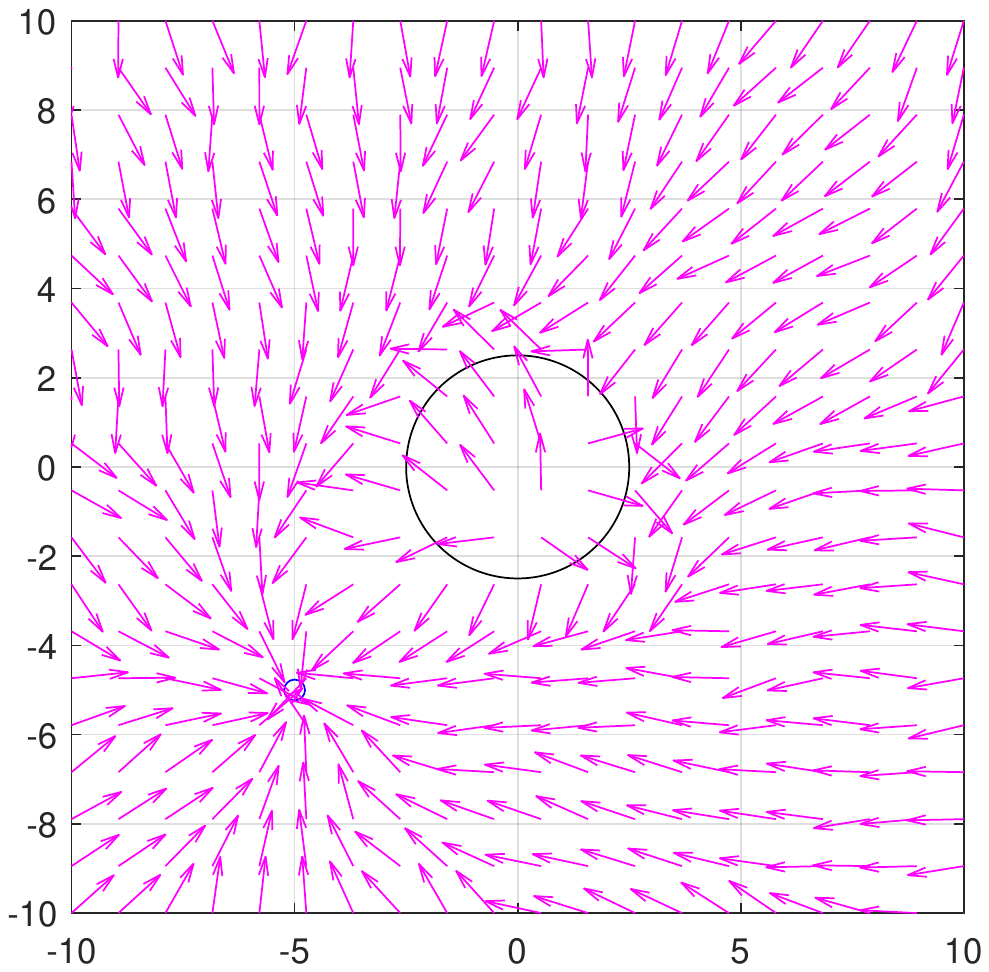} \\
		~&~&\\
		\small (a) & \small (b) & \small (c) 
	\end{tabular}
	\caption{Sample trajectories with fixed length $T = 100$ of learned policies corresponding to noiseless ((a): $N = 1$, $\omega = 0$) and noisy ((b): $N = 10$, $\omega =0.1$) settings (c) Learned field plot for ZDPG-S with noise level $\omega = 0.1$ shows that the variance in Figure \ref{fig:LC} comes only from the system noise.}
	\label{fig:traj}
	\vspace{-12pt}
\end{figure*}
Depending on the level of the system noise, ZDPG(-S) consistently outperforms PG-B (and also PG), both in mean performance and the associated variance. 
Interestingly, the improvement of ZDPG over PG-B is sharper when the agent dynamics are noisy (also see our Supplementary Material, Section \ref{supp:results}).
In fact, we observe that, either with \textit{or} without system noise, the perturbation of the initial state with a deterministic policy indeed provides more meaningful information for policy improvement than in the setting of a stochastic policy (also see Figure \ref{fig:traj}). 
One should also note that PG-B performs better in the noisy setting with MC variance reduction (\ref{fig:LC} (c)) than the noiseless setting (\ref{fig:LC} (a)). Indeed, variance reduction improves the performance of all methods, except for vanilla policy gradient. We emphasize though that we evaluate stochastic policies using only their means. Our empirical results corroborate the claim that randomness in the policy is \emph{not} necessary to solve MDPs within the PG-based RL framework.

\section{Conclusions and Future Work}
In this work, we introduced the ZDPG algorithm, which is based on an efficient two-point zeroth-order approximation of the deterministic policy gradient, for the infinite horizon Markov decision problem. Using random horizons to evaluate the zeroth-order estimates of the state-action function, we showed that low-dimensional perturbations in the action space can effectively replace the need for randomized policies in the classical policy gradient setting. We also presented finite sample complexity results for ZDPG, which improve upon the state of the art by up to two orders of magnitude. Finally, we corroborated our results with a numerical evaluation on a navigation problem which showcased the advantages of ZDPG over both standard PG and PG with a baseline. For future work, variance reduced ZDPG could be realized with more sample efficient variance reduction techniques, such as gradient averaging or using a critic network to evaluate zeroth order state action function.

%% file: appendix1.tex
\section*{Supplementary Material}

\section{Proof of Lemma \ref{lem:Bounded_Q_var}}
	%
	
Since the estimate $\hat{Q}$ is unbiased, it suffices to show that
$\mathbb{E}[|\hat{Q}|^{2}]$ is bounded. Indeed, we have
\begin{flalign}
\mathbb{E}[|\hat{Q}|^{2}] 
=\mathbb{E}\Bigg[\Bigg|\sum_{t=0}^{T}R(s_{t},a_{t}=\pi_{\theta}(s_{t}))\Bigg|^{2}\Bigg]\nonumber 
 & \le\mathbb{E}\Bigg[\Bigg(\sum_{t=0}^{T}|R(s_{t},a_{t}=\pi_{\theta}(s_{t}))|\Bigg)^{2}\Bigg]\nonumber \\
 & \le\mathbb{E}\big[(T+1)^{2}U_{R}^{2}\big]
 =U_{R}^{2}\dfrac{1+\gamma}{(1-\gamma)^{2}}, \nonumber
\end{flalign}
where the last equality follows from the fact that $T$ is a geometric
random variable with probability of success $1-\gamma$. Enough said.	\null\nobreak\hfill\ensuremath{\square}


\section{Proof of Theorem \ref{thm:nesterov_result1}}

	Let Assumptions \ref{asm:bounded_gradient}-\ref{asm:smooth} and also Assumption \ref{asm:nesterov_assumptions} be in effect. Also, recall the definitions of the gradients, namely
	\begin{equation} \nonumber
	\nabla J(\theta) = \E_{s\sim \rho^{\pi_\theta}}\left[\nabla_\theta \pi_\theta(s) \nabla_a Q^{\pi_\theta}(s,a)|_{a = \pi_{\theta}(s)}\right], 
	\end{equation}
	and 
	\begin{equation} \nonumber
	 \hat \nabla J(\theta) =  \E_{s\sim \rho^{\pi_\theta}} \left[\nabla_\theta \pi_\theta(s) \nabla_a Q_\mu^{\pi_\theta} (s, a) \vert_{a = \pi_\theta(s)  } \right].
	\end{equation}
	By definition, and then by Jensen, we have 
	\begin{equation}
	\begin{split}
	\| \nabla J(\theta) &-  \hat \nabla J(\theta)\| \\ \nonumber
	&\hspace{-24pt}= \Big\|
	\E_{s\sim \rho^{\pi_\theta}}\left[\nabla_\theta \pi_\theta(s) \nabla_a Q^{\pi_\theta}(s,a)\big|_{a = \pi_{\theta}(s)}\right] - \E_{s\sim \rho^{\pi_\theta}}\left[\nabla_\theta \pi_\theta(s) \nabla_a Q_\mu^{\pi_\theta}(s,a)\big|_{a = \pi_{\theta}(s)}\right] \Big\| \\ \nonumber
	&\hspace{-24pt}= \Big\| \E_{s\sim \rho^{\pi_\theta}}\left[\nabla_\theta \pi_\theta(s) \nabla_a Q^{\pi_\theta}(s,a)\big|_{a = \pi_{\theta}(s)} -\nabla_\theta \pi_\theta(s) \nabla_a Q_\mu^{\pi_\theta}(s,a)\big|_{a = \pi_{\theta}(s)}\right] \Big\|\\ \nonumber
	&\hspace{-24pt} \leq  \E_{s\sim \rho^{\pi_\theta}}\left[ \big\| \nabla_\theta \pi_\theta(s) \nabla_a Q^{\pi_\theta}(s,a)\big|_{a = \pi_{\theta}(s)} -\nabla_\theta \pi_\theta(s) \nabla_a Q_\mu^{\pi_\theta}(s,a)\big|_{a = \pi_{\theta}(s)}\big\|\right].
	\end{split}
	\end{equation}
	Grouping terms, then applying Cauchy-Schwarz, we obtain
	\begin{equation}
	\begin{split} \nonumber
	\| \nabla J(\theta) - \hat \nabla J(\theta)\| &\leq \E_{s\sim \rho^{\pi_\theta}}\left[ \big\| \nabla_\theta \pi_\theta(s)\big( \nabla_a Q^{\pi_\theta}(s,a)\big|_{a = \pi_{\theta}(s)} - \nabla_a Q_\mu^{\pi_\theta}(s,a)\big|_{a = \pi_{\theta}(s)}\big)\big\|\right] \\
	&\leq \E_{s\sim \rho^{\pi_\theta}}\left[ \left\| \nabla_\theta \pi_\theta(s) \right\| \cdot \left\| \nabla_a Q^{\pi_\theta}(s,a)\big|_{a = \pi_{\theta}(s)} - \nabla_a Q_\mu^{\pi_\theta}(s,a)\big|_{a = \pi_{\theta}(s)}\right\|\right]. \\
	\end{split}
	\end{equation}
	By Assumption \ref{asm:bounded_gradient}, we can bound $\nabla_\theta \pi_\theta(s)$ by $B_\Theta$. As such, the remaining expression can be bounded by invoking Assumption \ref{asm:nesterov_assumptions}, yielding 
	%
	\begin{equation}
	\begin{split} \nonumber
	\| \nabla J(\theta) - \hat \nabla J(\theta)\| & \leq B_\Theta\E_{s\sim \rho^{\pi_\theta}}\left[ \left\| \nabla_a \left( Q^{\pi_\theta}(s,a) -  Q_\mu^{\pi_\theta}(s,a)\right)\big|_{a = \pi_{\theta}(s)}\right\|\right] \\
	&\leq B_\Theta\E_{s\sim \rho^{\pi_\theta}}\left[ \left\| \nabla_a \left( Q^{\pi_\theta}(s,a) -   \E_\mathbf{u} \left[Q^{\pi_\theta}(s,a+ \mu \textbf{u})\right]\right)\big|_{a = \pi_{\theta}(s)}\right\|\right]\\
	&\leq  B_\Theta\E_{s\sim \rho^{\pi_\theta}}\left[ \left\| \nabla_a \left(\E_\mathbf{u} \left[ Q^{\pi_\theta}(s,a) -   Q^{\pi_\theta}(s,a+ \mu \textbf{u})\right]\right)\big|_{a = \pi_{\theta}(s)}\right\|\right]\\
	&\leq  B_\Theta\E_{s\sim \rho^{\pi_\theta}}\left[ \left\| \E_\mathbf{u} \left[\nabla_a \left( Q^{\pi_\theta}(s,a) -   Q^{\pi_\theta}(s,a+ \mu \textbf{u})\right]\right)\big|_{a = \pi_{\theta}(s)}\right\|\right]\\
	&\leq  B_\Theta\E_{s\sim \rho^{\pi_\theta}}\left[  \E_\mathbf{u} \left[\left\|\nabla_a \left( Q^{\pi_\theta}(s,a) -   Q^{\pi_\theta}(s,a+ \mu \textbf{u})\right)\big|_{a = \pi_{\theta}(s)}\right\|\right]\right]\\
	&\leq  B_\Theta\E_{s\sim \rho^{\pi_\theta}}\left[  \E_\mathbf{u} \left[G\left\|\mu \mathbf{u}\right\|\right]\right]\\
	&\leq B_\Theta \mu \sqrt{p} G/(1-\gamma),
	\end{split}
	\end{equation}
	where the final inequality comes from \cite[Lemma 1]{nesterov2017random}.

    If, alternatively, Assumption \ref{asm:nesterov_assumptions_H} instead of Assumption \ref{asm:nesterov_assumptions}, holds, then by \cite[Lemma 3]{nesterov2017random}, we obtain
	\begin{equation}
	\begin{split} \nonumber
	\| \nabla J(\theta) - \hat \nabla J(\theta)\| & \leq B_\Theta\E_{s\sim \rho^{\pi_\theta}}\left[ \left\| \nabla_a \left( Q^{\pi_\theta}(s,a) -  Q_\mu^{\pi_\theta}(s,a)\right)\big|_{a = \pi_{\theta}(s)}\right\|\right] \\
	&\leq B_\Theta \mu^2 (p+4)^2 H/(1-\gamma).
	\end{split}
	\end{equation}
    This completes the proof.\null\nobreak\hfill\ensuremath{\square}

\section{Proof of Lemma \ref{lem:bounded_variance}}

First, by definition, we may write
\begin{equation} \nonumber
(1-\gamma)^2\mathbb{E}[\Vert g_{t}\Vert^{2}]=\mathbb{E}\Bigg[\Bigg\Vert\nabla_{\theta}\pi_{\theta_{t}}(s_{t})\dfrac{\hat{Q}_{t}^{+}-\hat{Q}_{t}^{-}}{\mu}{\bf u}\Bigg\Vert^{2}\Bigg]=:\mathbb{E}[\Vert g'_{t}\Vert^{2}],
\end{equation}
where
\begin{align}\nonumber
\hat{Q}_{t}^{\pm} & =Q^{\pi_{\theta_{t}}}(\pm)+\dfrac{1}{N}\sum_{n=1}^{N}w_{t,n}^{\pm}.
\end{align}
By Cauchy-Schwarz, and by exploiting the above, we readily get
\begin{flalign}
\mathbb{E}[\Vert g'_{t}\Vert^{2}] & \le B_{\Theta}^{2}\mathbb{E}\Bigg[\Bigg\Vert\dfrac{Q^{\pi_{\theta_{t}}}(+)-Q^{\pi_{\theta_{t}}}(-)}{\mu}+\dfrac{1}{N}\sum_{n=1}^{N}\dfrac{w_{t,n}^{+}-w_{t,n}^{-}}{\mu}\Bigg\Vert^{2}\Vert{\bf u}\Vert^{2}\Bigg]\nonumber \\
 & =B_{\Theta}^{2}\mathbb{E}\Bigg[\Bigg|\dfrac{Q^{\pi_{\theta_{t}}}(s_{t},\pi_{\theta_{t}}(s_{t})+\mu{\bf u})-Q^{\pi_{\theta_{t}}}(s_{t},\pi_{\theta_{t}}(s_{t}))}{\mu}+\dfrac{1}{N}\sum_{n=1}^{N}\dfrac{w_{t,n}^{+}-w_{t,n}^{-}}{\mu}\Bigg|^{2}\Vert{\bf u}\Vert^{2}\Bigg]\nonumber \\
 & \le2B_{\Theta}^{2}\mathbb{E}\Bigg[\Bigg|\dfrac{Q^{\pi_{\theta_{t}}}(s_{t},\pi_{\theta_{t}}(s_{t})+\mu{\bf u})-Q^{\pi_{\theta_{t}}}(s_{t},\pi_{\theta_{t}}(s_{t}))}{\mu}\Bigg|^{2}\Vert{\bf u}\Vert^{2} \nonumber \\
 &\quad\quad\quad\quad\quad\quad\quad\quad\quad\quad\quad\quad\quad\quad\quad\quad\quad\quad+\Bigg|\dfrac{1}{N}\sum_{n=1}^{N}\dfrac{w_{t,n}^{+}-w_{t,n}^{-}}{\mu}\Bigg|^{2}\Vert{\bf u}\Vert^{2}\Bigg].\label{eq:Bound_1}
\end{flalign}
For the first term on the right-hand side of (\ref{eq:Bound_1}),
we have 
\begin{equation} 
\mathbb{E}\Bigg[\Bigg|\dfrac{Q^{\pi_{\theta_{t}}}(s_{t},\pi_{\theta_{t}}(s_{t})+\mu{\bf u})-Q^{\pi_{\theta_{t}}}(s_{t},\pi_{\theta_{t}}(s_{t}))}{\mu}\Bigg|^{2}\Vert{\bf u}\Vert^{2}\Bigg]\le L^2\mathbb{E}\big[\Vert{\bf u}\Vert^{4}\big]\le L^2 (p+4)^{2}.\label{eq:Bound_2}
\end{equation}
For the second term on the right-hand side of (\ref{eq:Bound_1}),
we may write
\begin{align} \nonumber
\mathbb{E}\Bigg[\Bigg|\dfrac{1}{N}\sum_{n=1}^{N}\dfrac{w_{t,n}^{+}-w_{t,n}^{-}}{\mu}\Bigg|^{2}\Vert{\bf u}\Vert^{2}\Bigg] & =\dfrac{1}{\mu^{2}N^{2}}\mathbb{E}\Bigg[\mathbb{E}\Bigg[\Bigg|\sum_{n=1}^{N}w_{t,n}^{+}-w_{t,n}^{-}\Bigg|^{2}\Vert{\bf u}\Vert^{2}\Bigg|s_t,{\bf u},\theta_{t},T_{Q}\Bigg]\hspace*{-1bp}\hspace*{-1bp}\Bigg]\nonumber \\
 & =\dfrac{1}{\mu^{2}N^{2}}\mathbb{E}\Bigg[\Vert{\bf u}\Vert^{2}\mathbb{E}\Bigg[\Bigg|\sum_{n=1}^{N}w_{t,n}^{+}-w_{t,n}^{-}\Bigg|^{2}\Bigg|s_t,{\bf u},\theta_{t},T_{Q}\Bigg]\hspace*{-1bp}\hspace*{-1bp}\Bigg]. \nonumber
\end{align}
Naturally, we need to focus on the (inner) conditional expectation
above. By construction, it is true that, for every $t$, the differences
$w_{t,n}^{+}-w_{t,n}^{-},n\in\mathbb{N}_{N}^{+}$ are \textit{conditionally}
independent and identically distributed relative to $s,{\bf u},\theta_{t}$
and $T_{Q}$, and \textit{with zero mean}. This implies that
\begin{flalign}
\mathbb{E}\Bigg[\Bigg|\sum_{n=1}^{N}w_{t,n}^{+}-w_{t,n}^{-}\Bigg|^{2}\Bigg|s_t,{\bf u},\theta_{t},T_{Q}\Bigg] & =\sum_{n=1}^{N}\mathbb{E}\big[\big|w_{t,n}^{+}-w_{t,n}^{-}\big|^{2}\big|s_t,{\bf u},\theta_{t},T_{Q}\big]\nonumber \\
 & =N\mathbb{E}\big[\big|w_{t,1}^{+}-w_{t,1}^{-}\big|^{2}\big|s_t,{\bf u},\theta_{t},T_{Q}\big],\nonumber \\
 & \le N2\mathbb{E}\big[\big(w_{t,1}^{+}\big)^{2}+\big(w_{t,1}^{-}\big)^{2}\big|s_t,{\bf u},\theta_{t},T_{Q}\big]\nonumber \\
 & \le N4U_{R}^{2}\dfrac{1+\gamma}{(1-\gamma)^{2}}=:N\sigma^{2},\nonumber
\end{flalign}
where in the last line we have used Lemma \ref{lem:Bounded_Q_var} twice. As a result, we obtain that
\begin{equation} 
\mathbb{E}\Bigg[\Bigg|\dfrac{1}{N}\sum_{n=1}^{N}\dfrac{w_{t,n}^{+}-w_{t,n}^{-}}{\mu}\Bigg|^{2}\Vert{\bf u}\Vert^{2}\Bigg]\le\dfrac{\sigma^{2}p}{\mu^{2}N},\label{eq:Bound_3}
\end{equation}
since $\mathbb{E}\big[\Vert{\bf u}\Vert^{2}\big]\equiv p$. Combining
(\ref{eq:Bound_2}) and (\ref{eq:Bound_3}), we end up with the uniform
bound
\begin{equation} \nonumber
\mathbb{E}[\Vert g_{t}\Vert^{2}]\le
\dfrac{2B_{\Theta}^{2}}{(1-\gamma)^2}
\Big(L^2(p+4)^{2}+\dfrac{\sigma^{2}p}{\mu^{2}N}\Big),
\end{equation}
thus completing the proof.
\null\nobreak\hfill\ensuremath{\square}
\section{Proof of Theorem \ref{thm:main_1}}

First, it is not hard to see that the stochastic ascend direction
$g_{t}$ constructed by Algorithm \ref{alg:ZOPG} is an unbiased estimate of the
quasi-gradient $\hat{\nabla}J$. Indeed, with $s_{t}\sim (1-\gamma)\rho^{\pi_{\theta_{t}}}$,
we may carefully write
\begin{flalign}
(1-\gamma)\mathbb{E}[g_{t}|\theta_{t}] & =\mathbb{E}\Bigg[\nabla_{\theta}\pi_{\theta_{t}}(s_{t})\dfrac{\hat{Q}_{t}^{+}-\hat{Q}_{t}^{-}}{\mu}{\bf u}\Bigg|\theta_{t}\Bigg]\nonumber \\
 & =\mathbb{E}\Bigg[\mathbb{E}\Bigg[\nabla_{\theta}\pi_{\theta_{t}}(s_{t})\dfrac{\hat{Q}_{t}^{+}(s_{t},{\bf u},\theta_{t},T_{Q})-\hat{Q}_{t}^{-}(s_{t},\theta_{t},T_{Q})}{\mu}{\bf u}\Bigg|s_{t},{\bf u},\theta_{t},T_{Q}\Bigg]\hspace*{-1bp}\hspace*{-1bp}\Bigg|\theta_{t}\Bigg]\nonumber \\
 & \equiv\mathbb{E}\Bigg[\nabla_{\theta}\pi_{\theta_{t}}(s_{t})\mathbb{E}\Bigg[\dfrac{\hat{Q}_{t}^{+}(s_{t},{\bf u},\theta_{t},T_{Q})-\hat{Q}_{t}^{-}(s_{t},\theta_{t},T_{Q})}{\mu}\Bigg|s_{t},{\bf u},\theta_{t},T_{Q}\Bigg]\hspace*{-1bp}{\bf u}\Bigg|\theta_{t}\Bigg]\nonumber \\
 & =\mathbb{E}\Bigg[\nabla_{\theta}\pi_{\theta_{t}}(s_{t})\dfrac{Q^{\pi_{\theta_{t}}}(s_{t},\pi_{\theta_{t}}(s_{t})+\mu{\bf u})-Q^{\pi_{\theta_{t}}}(s_{t},\pi_{\theta_{t}}(s_{t}))}{\mu}{\bf u}\Bigg|\theta_{t}\Bigg]\nonumber \\
 & =\mathbb{E}\Bigg[\nabla_{\theta}\pi_{\theta_{t}}(s_{t})\mathbb{E}\Bigg[\dfrac{Q^{\pi_{\theta_{t}}}(s_{t},\pi_{\theta_{t}}(s_{t})+\mu{\bf u})-Q^{\pi_{\theta_{t}}}(s_{t},\pi_{\theta_{t}}(s_{t}))}{\mu}{\bf u}\Bigg|s_{t},\theta_{t}\Bigg]\hspace*{-1bp}\hspace*{-1bp}\Bigg|\theta_{t}\Bigg]\nonumber \\
 & =\mathbb{E}\big[\nabla_{\theta}\pi_{\theta_{t}}(s_{t})\nabla_{a}Q_{\mu}^{\pi_{\theta_{t}}}(s,a)|_{a=\pi_{\theta_{t}}(s_{t})}|\theta_{t}\big]\nonumber \\
 & = 
 (1-\gamma)
 \mathbb{E}_{s_{t}\sim\rho^{\pi_{\theta_{t}}}}\big[\nabla_{\theta}\pi_{\theta_{t}}(s_{t})\nabla_{a}Q_{\mu}^{\pi_{\theta_{t}}}(s,a)|_{a=\pi_{\theta_{t}}(s_{t})}|\theta_{t}\big] \nonumber \\
 & = 
 (1-\gamma)
 \hat{\nabla}J(\theta_t). \nonumber
\end{flalign}

Next, using the assumption that $J(\theta)$ is $L_J$-smooth, we have
\begin{equation} \nonumber
    J(\theta_{t + 1}) \geq J(\theta_t) + (\theta_{t+1} - \theta_t)^\top \nabla J(\theta_t) - L_J\|\theta_{t+1} - \theta_t\|^2
\end{equation}
From Algorithm \ref{alg:ZOPG}, we have $\theta_{t+1} - \theta_t = \alpha g_t$. Let $\E_t$ denote conditional expectation taken relative to history up to iteration $t-1$. Then, by taking the expectation on both sides, we obtain
\begin{equation} \nonumber
    \E_t\left[J(\theta_{t+1})\right] \geq J(\theta_t) + \alpha \hat \nabla J(\theta_t)^\top \nabla J(\theta_t) - L_J\alpha^2 \E_t\left[ \| g_t  \|^2 \right].
\end{equation}
Now add and subtract $\alpha  \nabla J(\theta_t)^\top \nabla J(\theta_t)$ to the right hand side. By grouping terms, we may write
\begin{equation} \nonumber
    \E_t  \left[J(\theta_{t+1})\right] \ge 
    J(\theta_t) + \alpha  \big( \hat \nabla J(\theta_t) - \nabla J(\theta_t)\big)^{\hspace{-1pt}\top} \nabla J(\theta_t) + \alpha  \|\nabla J(\theta_t)\|^2 - L_J\alpha^2\E_t\left[ \| g_t  \|^2 \right].
\end{equation}
Using Cauchy-Schwarz, we bound the term $\big( \hat \nabla J(\theta_t) - \nabla J(\theta_t)\big)^\top \nabla J(\theta_t)$ by $-\| \hat \nabla J(\theta_t) - \nabla J(\theta_t)\|\cdot \|\nabla J(\theta_t)\|$. Further, because $J(\theta)$ is assumed to be $G_J$-Lipschitz, we have
\begin{equation} \nonumber
\E_t \left[J(\theta_{t+1})\right] \geq J(\theta_t) - \alpha  G_J \|\hat \nabla J(\theta_t) - \nabla  J(\theta_t)\| + \alpha  \|\nabla J(\theta_t)\|^2 - L_J\alpha^2 \E_t\left[ \| g_t  \|^2 \right].
\end{equation}
Taking the total expectation, invoking Lemma \ref{lem:bounded_variance}, summing up until $T$, and dividing by $1/(\alpha T)$, we obtain
\vspace{-11pt}
\begin{equation} 
\begin{split}\label{equ:before_rearranging} 
 \frac{1}{T} \sum_{t = 1}^T \E  \|\nabla J(\theta_t)\|^2 \leq & \hspace{1pt} \frac{\Delta_0}{\alpha T}   
 +  L_J\alpha \frac{2B_\Theta^2}{(1-\gamma)^2}(p+4)^2L^2 
 + L_J\alpha \frac{2B_\Theta^2}{(1-\gamma)^2}\frac{p \sigma^2}{\mu^2N}  \\
 &+ \frac{ G_J}{T} \sum_{t = 1}^T \E [\|\hat \nabla J(\theta_t) - \nabla J(\theta_t)\|],
 \end{split}
\end{equation}
where $\Delta_0 =  J(\theta^*)-J(\theta_0) $ and $\theta^* \in \arg\max J(\theta)$.

Under Assumption \ref{asm:nesterov_assumptions}, Theorem \ref{thm:nesterov_result1} implies that
\begin{equation} 
\frac{1}{T} \sum_{t = 1}^T \E  \|\nabla J(\theta_t)\|^2 \leq \frac{\Delta_0}{\alpha T}  
+  L_J\alpha \frac{2B_\Theta^2}{(1-\gamma)^2}(p+4)^2L^2 
 + L_J\alpha \frac{2B_\Theta^2}{(1-\gamma)^2}\frac{p \sigma^2}{\mu^2N}
+ G_J\frac{B_\Theta \mu \sqrt{p}G}{1-\gamma}. \nonumber
\end{equation}
If, alternatively, Assumption \ref{asm:nesterov_assumptions_H} holds, then from \eqref{equ:before_rearranging} we use the pertinent result form Theorem \ref{thm:nesterov_result1} to obtain
\begin{equation} 
\frac{1}{T} \sum_{t = 1}^T \E  \|\nabla J(\theta_t)\|^2 \leq \frac{\Delta_0}{\alpha T}  
+  L_J\alpha \frac{2B_\Theta^2}{(1-\gamma)^2}(p+4)^2L^2
 + L_J\alpha \frac{2B_\Theta^2}{(1-\gamma)^2}\frac{p \sigma^2}{\mu^2N}
+ G_J \frac{B_\Theta \mu^2 (p+4)^2H}{1-\gamma}. \nonumber
\end{equation}
To complete the proof, we set the number of Monte Carlo rollouts ($N$), smoothing parameter ($\mu$), and step size ($\alpha$) in 
each of the expressions above, according to Table \ref{sample-table}. Enough said.
\null\nobreak\hfill\ensuremath{\square}

\textbf{Extensions for constrained policy parameterizations:} In such a case, we may consider a projected gradient step in Algorithm \ref{alg:ZOPG}, i.e., 
\begin{equation}
\theta_{t+1}=\Pi_{\Theta}\{\theta_{t}+\alpha g_{t}\},\nonumber
\end{equation}
where $\Theta\subset\mathbb{R}^{d}$ is also assumed to be closed
and convex. Below we present two approaches for deriving meaningful convergence rates.

\textit{Noisy projection approach:} In this approach, we may expand the product $(\theta_{t+1}-\theta_{t})^{\top}\nabla J(\theta_{t})$
into two relevant terms as
\begin{align*}
(\theta_{t+1}-\theta_{t})^{\top}\nabla J(\theta_{t}) & =(\theta_{t+1}+\Pi_{\Theta}\{\theta_{t}+\alpha\nabla J(\theta_{t})\}-\Pi_{\Theta}\{\theta_{t}+\alpha\nabla J(\theta_{t})\}-\theta_{t})^{\top}\nabla J(\theta_{t})\\
 & =(\Pi_{\Theta}\{\theta_{t}+\alpha\nabla J(\theta_{t})\}-\theta_{t})^{\top}\nabla J(\theta_{t})\\
 &\quad\quad+(\theta_{t+1}-\Pi_{\Theta}\{\theta_{t}+\alpha\nabla J(\theta_{t})\})^{\top}\nabla J(\theta_{t}).
\end{align*}
For the first term, we invoke the key property \cite[Lemma 1]{ghadimi2016mini}
\begin{equation}
(\Pi_{\Theta}\{\theta+\alpha g\}-\theta)^{\top}g\ge\dfrac{1}{\alpha}\Vert\Pi_{\Theta}\{\theta+\alpha g\}-\theta\Vert^{2},\quad\forall(\theta,g)\in\Theta\times\mathbb{R}^{d}, \label{GenProj}
\end{equation}
yielding
\[
(\Pi_{\Theta}\{\theta_{t}+\alpha\nabla J(\theta_{t})\}-\theta_{t})^{\top}\nabla J(\theta_{t})\ge\dfrac{1}{\alpha}\Vert\theta_{t}-\Pi_{\Theta}\{\theta_{t}+\alpha\nabla J(\theta_{t})\}\Vert^{2}.
\]
For the second term, we may further expand as
\begin{flalign*}
 & \hspace*{-1bp}\hspace*{-1bp}\hspace*{-1bp}\hspace*{-1bp}\hspace*{-1bp}\hspace*{-1bp}\hspace*{-1bp}\hspace*{-1bp}\hspace*{-1bp}\hspace*{-1bp}\hspace*{-1bp}\hspace*{-1bp}\hspace*{-1bp}\hspace*{-1bp}\hspace*{-1bp}\hspace*{-1bp}\hspace*{-1bp}\hspace*{-1bp}(\theta_{t+1}-\Pi_{\Theta}\{\theta_{t}+\alpha\nabla J(\theta_{t})\})^{\top}\nabla J(\theta_{t})\\
 & =(\theta_{t+1}-\Pi_{\Theta}\{\theta_{t}+\alpha\hat{\nabla}J(\theta_{t})\})^{\top}\nabla J(\theta_{t})\\
 & \quad\quad+(\Pi_{\Theta}\{\theta_{t}+\alpha\hat{\nabla}J(\theta_{t})\}-\Pi_{\Theta}\{\theta_{t}+\alpha\nabla J(\theta_{t})\})^{\top}\nabla J(\theta_{t}),
\end{flalign*}
which implies that
\begin{flalign*}
 & \hspace*{-1bp}\hspace*{-1bp}\hspace*{-1bp}\hspace*{-1bp}\hspace*{-1bp}\hspace*{-1bp}\hspace*{-1bp}\hspace*{-1bp}\hspace*{-1bp}\hspace*{-1bp}\hspace*{-1bp}\hspace*{-1bp}\hspace*{-1bp}\hspace*{-1bp}\hspace*{-1bp}\hspace*{-1bp}\hspace*{-1bp}\hspace*{-1bp}\mathbb{E}_{t}\{(\theta_{t+1}-\Pi_{\Theta}\{\theta_{t}+\alpha\nabla J(\theta_{t})\})^{\top}\nabla J(\theta_{t})\}\\
 & =(\mathbb{E}_{t}\{\theta_{t+1}\}-\Pi_{\Theta}\{\theta_{t}+\alpha\hat{\nabla}J(\theta_{t})\})^{\top}\nabla J(\theta_{t})\\
 & \quad\quad+(\Pi_{\Theta}\{\theta_{t}+\alpha\hat{\nabla}J(\theta_{t})\}-\Pi_{\Theta}\{\theta_{t}+\alpha\nabla J(\theta_{t})\})^{\top}\nabla J(\theta_{t})\\
 & \ge-\Vert\nabla J(\theta_{t})\Vert\Vert\mathbb{E}_{t}\{\theta_{t+1}\}-\Pi_{\Theta}\{\theta_{t}+\alpha\hat{\nabla}J(\theta_{t})\Vert\\
 & \quad\quad-\alpha\Vert\nabla J(\theta_{t})\Vert\Vert\hat{\nabla}J(\theta_{t})-\nabla J(\theta_{t})\Vert\\
 & \ge-G_{J}\Vert\mathbb{E}_{t}\{\theta_{t+1}\}-\Pi_{\Theta}\{\theta_{t}+\alpha\hat{\nabla}J(\theta_{t})\Vert\\
 & \quad\quad-\alpha G_{J}\Vert\hat{\nabla}J(\theta_{t})-\nabla J(\theta_{t})\Vert.
\end{flalign*}
Consequently, we obtain the lower bound
\begin{flalign*}
\mathbb{E}_{t}\{(\theta_{t+1}-\theta_{t})^{\top}\nabla J(\theta_{t})\} & \ge\dfrac{1}{\alpha}\Vert\theta_{t}-\Pi_{\Theta}\{\theta_{t}+\alpha\nabla J(\theta_{t})\}\Vert^{2}\\
 & \quad\;\,\,-\alpha G_{J}\Vert\hat{\nabla}J(\theta_{t})-\nabla J(\theta_{t})\Vert\\
 & \quad\quad-G_{J}\Vert\mathbb{E}_{t}\{\theta_{t+1}\}-\Pi_{\Theta}\{\theta_{t}+\alpha\hat{\nabla}J(\theta_{t})\Vert.
\end{flalign*}
Exploiting smoothness as in the unconstrained case discussed above,
we end up with the complexity estimate
\begin{flalign*}
 & \hspace*{-1bp}\hspace*{-1bp}\hspace*{-1bp}\hspace*{-1bp}\hspace*{-1bp}\hspace*{-1bp}\hspace*{-1bp}\hspace*{-1bp}\hspace*{-1bp}\hspace*{-1bp}\hspace*{-1bp}\hspace*{-1bp}\hspace*{-1bp}\hspace*{-1bp}\hspace*{-1bp}\dfrac{1}{T}\sum_{t=1}^{T}\mathbb{E}\dfrac{1}{\alpha^{2}}\Vert \Pi_{\Theta}\{\theta_{t}+\alpha\nabla J(\theta_{t})\} - \theta_{t}\Vert^{2}\\
 & \le\dfrac{\Delta_{0}}{\alpha T}
 +  L_J\alpha \frac{2B_\Theta^2}{(1-\gamma)^2}(p+4)^2L^2
 + L_J\alpha \frac{2B_\Theta^2}{(1-\gamma)^2}\frac{p \sigma^2}{\mu^2N}
 \\
 & \quad+\dfrac{G_{J}}{T}\sum_{t=1}^{T}\mathbb{E}\big[\Vert\hat{\nabla}J(\theta_{t})-\nabla J(\theta_{t})\Vert\big]
 \\ &  \quad\quad
 +\dfrac{G_{J}}{\alpha T}\sum_{t=1}^{T}\mathbb{E}\big[\Vert\mathbb{E}_{t}[\Pi_{\Theta}\{\theta_{t}+\alpha g_{t}\}]-\Pi_{\Theta}\{\theta_{t}+\alpha\hat{\nabla}J(\theta_{t})\}\Vert\big].
\end{flalign*}
Note that, in the above, whenever it is actually the case that $\theta_{t}+\alpha g_{t}\in\Theta$
\textit{conditioned on} $\theta_{t}$, it follows that (note that $\Theta$
is closed convex)
\begin{flalign*}
\Theta\ni\mathbb{E}_{t}[\Pi_{\Theta}\{\theta_{t}+\alpha g_{t}\}] & =\mathbb{E}_{t}[\theta_{t}+\alpha g_{t}]\\
 & =\theta_{t}+\alpha\hat{\nabla}J(\theta_{t}),
\end{flalign*}
which implies that
\[
\mathbb{E}_{t}[\Pi_{\Theta}\{\theta_{t}+\alpha g_{t}\}]-\Pi_{\Theta}\{\theta_{t}+\alpha\hat{\nabla}J(\theta_{t})\}=0.
\]
This will happen when $g_{t}$ is (or can be made) uniformly not too
large given $\theta_{t}$, and at the same time the stepsize
$\alpha$ is sufficiently small. Exactly such a behavior has been
observed in our simulations. 

\textit{Mini-batch approach:} Alternatively, we may proceed to derive a slightly more pessimistic complexity estimate by writing
\begin{align*}
(\theta_{t+1}-\theta_{t})^{\top}\nabla J(\theta_{t}) & =(\theta_{t+1}-\theta_{t})^{\top}(\nabla J(\theta_{t})-\hat{\nabla}J(\theta_{t})+\hat{\nabla}J(\theta_{t}))\\
 & =(\theta_{t+1}-\theta_{t})^{\top}(\nabla J(\theta_{t})-\hat{\nabla}J(\theta_{t}))+(\theta_{t+1}-\theta_{t})^{\top}\hat{\nabla}J(\theta_{t}).
\end{align*}
For the first term, Cauchy-Schwarz trivially gives
\[
(\theta_{t+1}-\theta_{t})^{\top}(\nabla J(\theta_{t})-\hat{\nabla}J(\theta_{t}))\ge-\alpha\Vert g_{t}\Vert\Vert\nabla J(\theta_{t})-\hat{\nabla}J(\theta_{t})\Vert.
\]
For the second term, we may further expand as
\begin{flalign*}
(\theta_{t+1}-\theta_{t})^{\top}\hat{\nabla}J(\theta_{t}) & =(\theta_{t+1}-\theta_{t})^{\top}(\hat{\nabla}J(\theta_{t})+g_{t}-g_{t})\\
 & =(\theta_{t+1}-\theta_{t})^{\top}(\hat{\nabla}J(\theta_{t})-g_{t})+(\theta_{t+1}-\theta_{t})^{\top}g_{t},
\end{flalign*}
where again invoking \eqref{GenProj} yields
\[
(\theta_{t+1}-\theta_{t})^{\top}g_{t}\ge\dfrac{1}{\alpha}\Vert\Pi_{\Theta}\{\theta_{t}+\alpha g_{t}\}-\theta_{t}\Vert^{2},
\]
whereas it also holds that
\begin{flalign*}
(\theta_{t+1}-\theta_{t})^{\top}(\hat{\nabla}J(\theta_{t})-g_{t}) & =(\Pi_{\Theta}\{\theta_{t}+\alpha g_{t}\}-\theta_{t})^{\top}(\hat{\nabla}J(\theta_{t})-g_{t})\\
 & =(\Pi_{\Theta}\{\theta_{t}+\alpha g_{t}\}-\Pi_{\Theta}\{\theta_{t}+\alpha\hat{\nabla}J(\theta_{t})\})^{\top}(\hat{\nabla}J(\theta_{t})-g_{t})\\
 & \quad\;\,\,+(\Pi_{\Theta}\{\theta_{t}+\alpha\hat{\nabla}J(\theta_{t})\}-\theta_{t})^{\top}(\hat{\nabla}J(\theta_{t})-g_{t}).
\end{flalign*}
In regard, to the latter expression, it is then true that
\begin{flalign*}
\mathbb{E}_{t}[(\theta_{t+1}-\theta_{t})^{\top}(\hat{\nabla}J(\theta_{t})-g_{t})] & =\mathbb{E}_{t}[(\Pi_{\Theta}\{\theta_{t}+\alpha g_{t}\}-\Pi_{\Theta}\{\theta_{t}+\alpha\hat{\nabla}J(\theta_{t})\})^{\top}(\hat{\nabla}J(\theta_{t})-g_{t})]\\
 & \quad\quad+0\\
 & \ge-\alpha\mathbb{E}_{t}[\Vert g_{t}-\hat{\nabla}J(\theta_{t})\Vert^{2}].
\end{flalign*}
Combining everything, we get
\begin{align*}
\mathbb{E}_{t}[(\theta_{t+1}-\theta_{t})^{\top}\nabla J(\theta_{t})] & \ge\dfrac{1}{\alpha}\Vert\Pi_{\Theta}\{\theta_{t}+\alpha g_{t}\}-\theta\Vert^{2}\\
 & \quad\;\,\,-\alpha\mathbb{E}_{t}[\Vert g_{t}\Vert]\Vert\nabla J(\theta_{t})-\hat{\nabla}J(\theta_{t})\Vert\\
 & \quad\quad\;\,\,-\alpha\mathbb{E}_{t}[\Vert g_{t}-\hat{\nabla}J(\theta_{t})\Vert^{2}].
\end{align*}
Now, note that that we can write
\begin{align*}
 & \hspace*{-1bp}\hspace*{-1bp}\hspace*{-1bp}\hspace*{-1bp}\hspace*{-1bp}\hspace*{-1bp}\hspace*{-1bp}\hspace*{-1bp}\hspace*{-1bp}\hspace*{-1bp}\hspace*{-1bp}\hspace*{-1bp}\dfrac{1}{\alpha^{2}}\mathbb{E}[\Vert\Pi_{\Theta}\{\theta_{t}+\alpha\nabla J(\theta_{t})\}-\theta_{t}\Vert^{2}]\\
 & =\dfrac{1}{\alpha^{2}}\mathbb{E}[\Vert\Pi_{\Theta}\{\theta_{t}+\alpha\nabla J(\theta_{t})\}-\Pi_{\Theta}\{\theta_{t}+\alpha\hat{\nabla}J(\theta_{t})\}\\
 & \quad\quad\quad\quad\quad\quad\quad\quad\quad\quad\quad+\Pi_{\Theta}\{\theta_{t}+\alpha\hat{\nabla}J(\theta_{t})\}-\theta_{t}\Vert^{2}]\\
 & \le2\mathbb{E}[\Vert\nabla J(\theta_{t})-\hat{\nabla}J(\theta_{t})\Vert^{2}]+2\dfrac{1}{\alpha^{2}}\mathbb{E}[\Vert\Pi_{\Theta}\{\theta_{t}+\alpha\hat{\nabla}J(\theta_{t})\}-\theta_{t}\Vert^{2}]\\
 & =2\mathbb{E}[\Vert\nabla J(\theta_{t})-\hat{\nabla}J(\theta_{t})\Vert^{2}]\\
 & \quad\;\,\,+2\dfrac{1}{\alpha^{2}}\mathbb{E}[\Vert\Pi_{\Theta}\{\theta_{t}+\alpha\hat{\nabla}J(\theta_{t})\}-\Pi_{\Theta}\{\theta_{t}+\alpha g_{t}\}\\
 & \quad\quad\quad\quad\quad\quad\quad\quad\quad\quad\quad\quad\;\,\,+\Pi_{\Theta}\{\theta_{t}+\alpha g_{t})\}-\theta_{t}\Vert^{2}]\\
 & \le2\mathbb{E}[\Vert\nabla J(\theta_{t})-\hat{\nabla}J(\theta_{t})\Vert^{2}]\\
 & \quad\;\,\,+4\mathbb{E}[\Vert g_{t}-\hat{\nabla}J(\theta_{t})\Vert^{2}]\\
 & \quad\quad\;\,\,+4\dfrac{1}{\alpha^{2}}\mathbb{E}[\Vert\Pi_{\Theta}\{\theta_{t}+\alpha g_{t}\}-\theta_{t}\Vert^{2}].
\end{align*}
Finally, following the usual steps as above, we obtain the complexity
estimate
\begin{flalign*}
 & \hspace*{-1bp}\hspace*{-1bp}\hspace*{-1bp}\hspace*{-1bp}\hspace*{-1bp}\hspace*{-1bp}\hspace*{-1bp}\hspace*{-1bp}\hspace*{-1bp}\hspace*{-1bp}\hspace*{-1bp}\hspace*{-1bp}\hspace*{-1bp}\hspace*{-1bp}\hspace*{-1bp}\dfrac{1}{T}\sum_{t=1}^{T}\mathbb{E}\dfrac{1}{\alpha^{2}}\Vert\Pi_{\Theta}\{\theta_{t}+\alpha\nabla J(\theta_{t})\}-\theta_{t}\Vert^{2}\\
 & \le\dfrac{2}{T}\sum_{t=1}^{T}\mathbb{E}[\Vert\nabla J(\theta_{t})-\hat{\nabla}J(\theta_{t})\Vert^{2}]+\dfrac{4}{T}\sum_{t=1}^{T}\mathbb{E}[\Vert g_{t}-\hat{\nabla}J(\theta_{t})\Vert^{2}]\\
 & \quad + \dfrac{\Delta_{0}}{\alpha T}
 +  L_J\alpha \frac{2B_\Theta^2}{(1-\gamma)^2}(p+4)^2L^2 
 + L_J\alpha \frac{2B_\Theta^2}{(1-\gamma)^2}\frac{p \sigma^2}{\mu^2N}
 \\
 & \quad\quad+\dfrac{1}{T}\sum_{t=1}^{T}\mathbb{E}[\Vert g_{t}\Vert\Vert\nabla J(\theta_{t})-\hat{\nabla}J(\theta_{t})\Vert]+\dfrac{1}{T}\sum_{t=1}^{T}\mathbb{E}[\Vert g_{t}-\hat{\nabla}J(\theta_{t})\Vert^{2}],
\end{flalign*}
or, more compactly,
\begin{flalign*}
 & \hspace*{-1bp}\hspace*{-1bp}\hspace*{-1bp}\hspace*{-1bp}\hspace*{-1bp}\hspace*{-1bp}\hspace*{-1bp}\hspace*{-1bp}\hspace*{-1bp}\hspace*{-1bp}\hspace*{-1bp}\hspace*{-1bp}\hspace*{-1bp}\hspace*{-1bp}\hspace*{-1bp}\dfrac{1}{T}\sum_{t=1}^{T}\mathbb{E}\dfrac{1}{\alpha^{2}}\Vert\Pi_{\Theta}\{\theta_{t}+\alpha\nabla J(\theta_{t})\}-\theta_{t}\Vert^{2}\\
 & \le\dfrac{\Delta_{0}}{\alpha T}
 +  L_J\alpha \frac{2B_\Theta^2}{(1-\gamma)^2}(p+4)^2L^2
 + L_J\alpha \frac{2B_\Theta^2}{(1-\gamma)^2}\frac{p \sigma^2}{\mu^2N}
 \\
 & \quad+\dfrac{1}{T}\sum_{t=1}^{T}\mathbb{E}[\Vert g_{t}\Vert\Vert\nabla J(\theta_{t})-\hat{\nabla}J(\theta_{t})\Vert]+\dfrac{2}{T}\sum_{t=1}^{T}\mathbb{E}[\Vert\nabla J(\theta_{t})-\hat{\nabla}J(\theta_{t})\Vert^{2}]\\
 & \quad\quad+\dfrac{5}{T}\sum_{t=1}^{T}\mathbb{E}[\Vert g_{t}-\hat{\nabla}J(\theta_{t})\Vert^{2}],
\end{flalign*}

This bound suggests that further variance reduction can be employed
in order to make the term involving $\mathbb{E}[\Vert g_{t}-\hat{\nabla}J(\theta_{t})\Vert^{2}]=\mathbb{E}[\Vert g_{t}-\mathbb{E}[g_{t}|\theta_{t}]\Vert^2]$
small, for instance by exploiting mini-batching, similar in spirit
to \cite{ghadimi2016mini}. However, this comes at the expense of significantly
higher training overhead.


\section{Complementary Numerical Results} \label{supp:results}

\begin{figure*}[!t]
	\centering
	\begin{tabular}{cc}
		\hspace{5.2mm}\includegraphics[trim = {5cm, 10cm, 5cm, 10cm}, width=.415\linewidth]{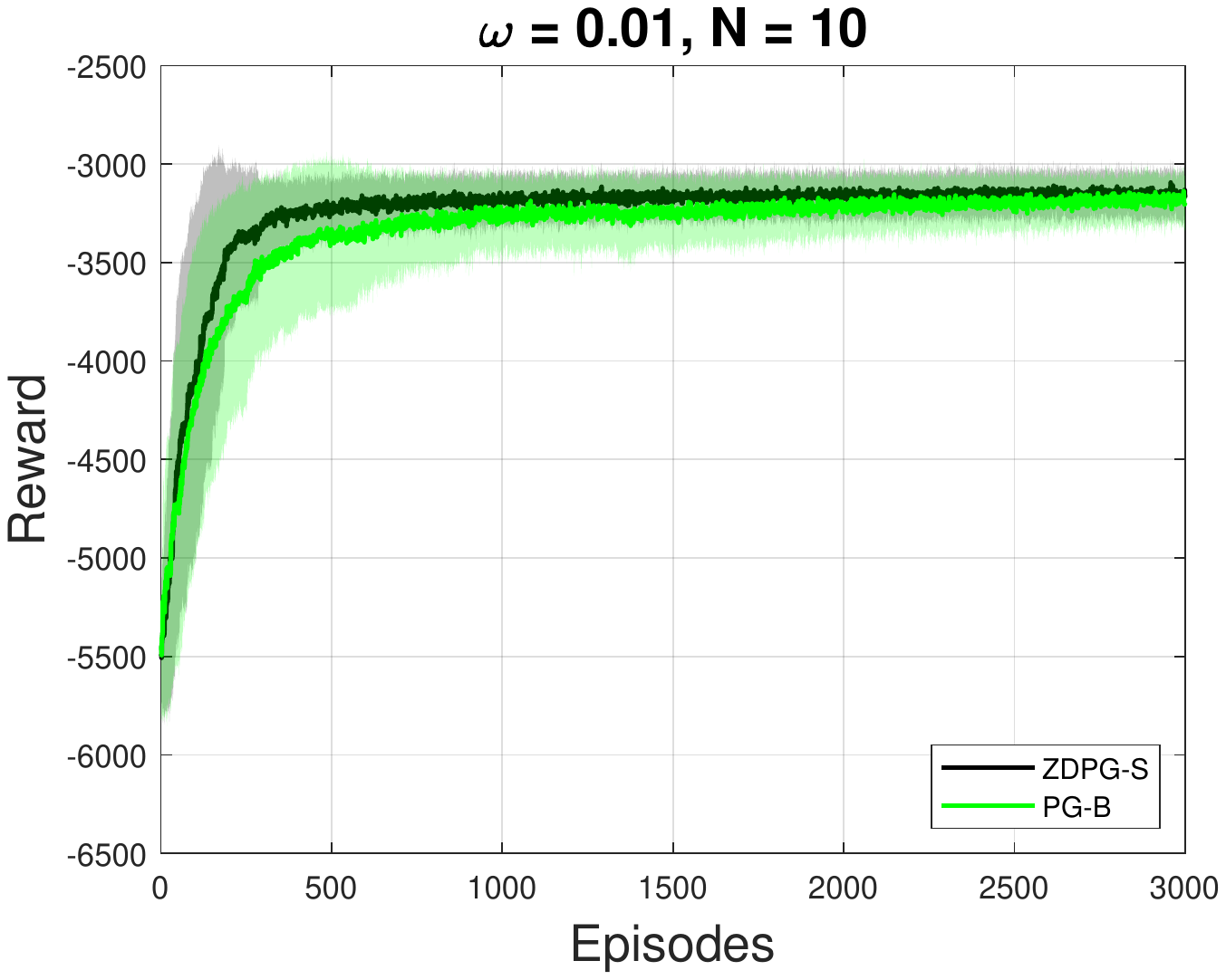}   &\hspace{8mm}
		\includegraphics[trim = {5cm, 10cm, 5cm, 10cm},width=.415\linewidth]{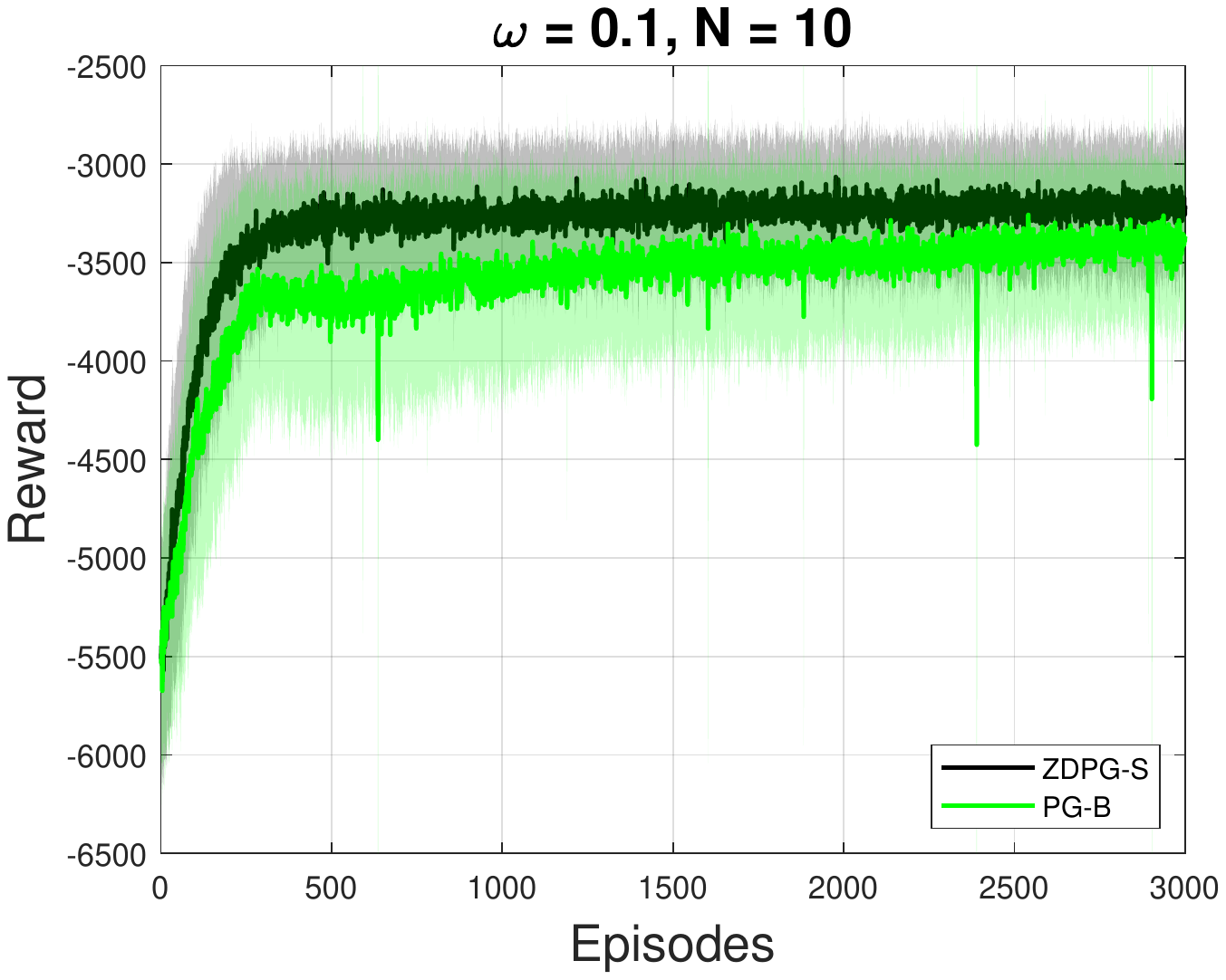} \\ \vspace{.5cm}
		~&~\\
		\hspace{5.2mm}\includegraphics[trim = {5cm, 10cm, 5cm, 10cm}, width=.415\linewidth]{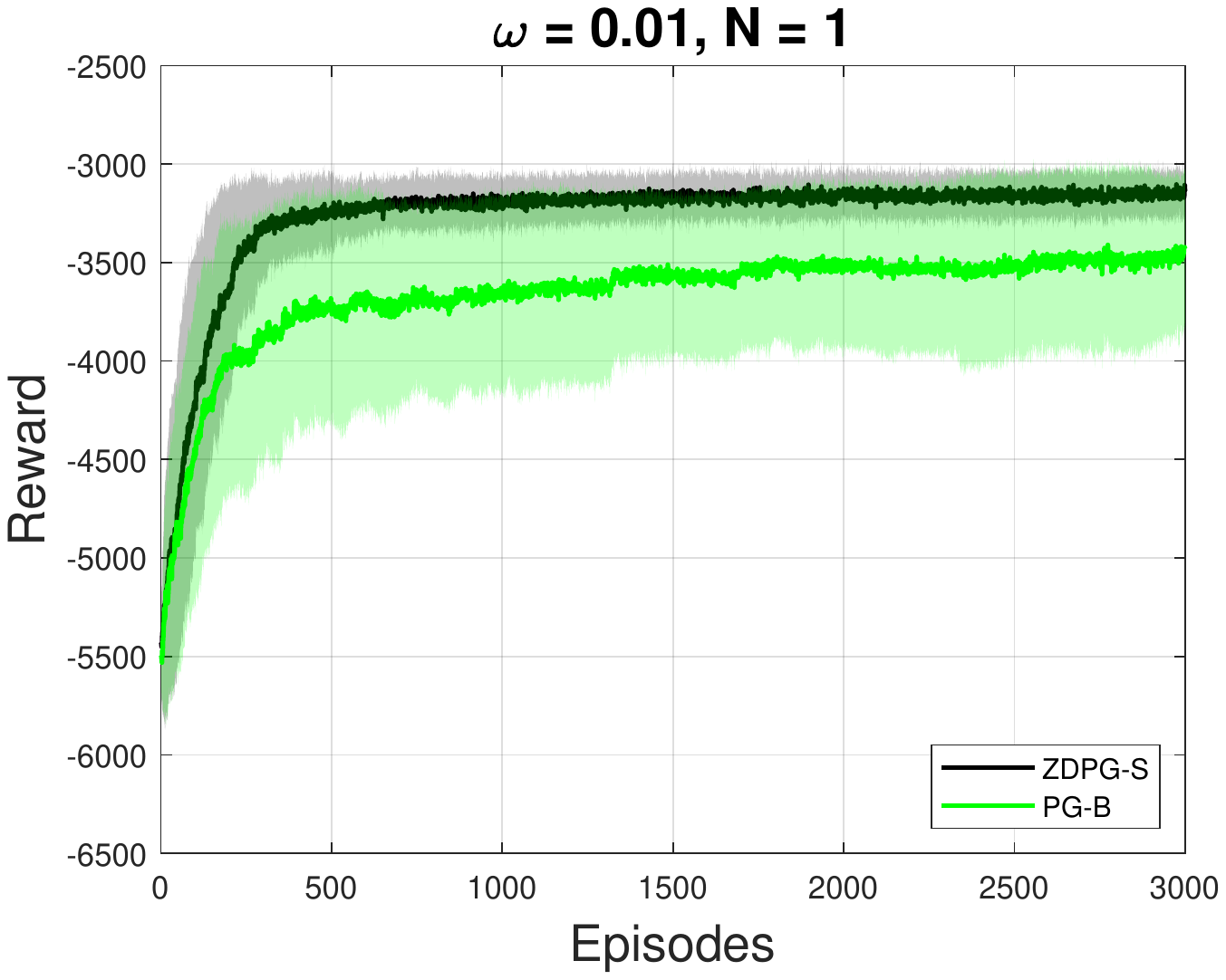}   &\hspace{8mm}
		\includegraphics[trim = {5cm, 10cm, 5cm, 10cm},width=.415\linewidth]{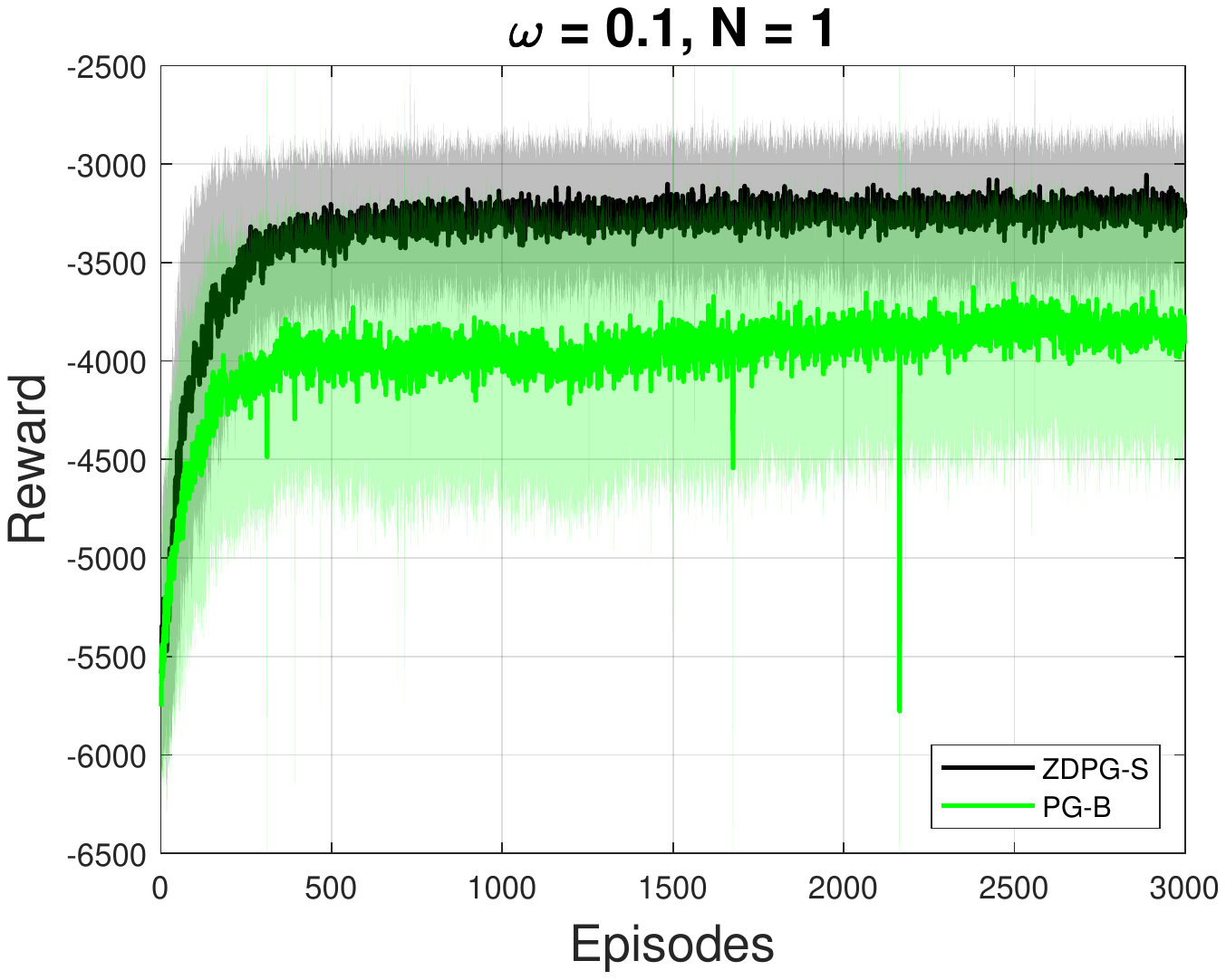} \\ \vspace{.4cm}
		~&~\\
	\end{tabular}
	\caption{Average reward per episode of length $T = 20$ with confidence bounds over 50 trials.  }\vspace{4mm}
	\label{fig:LC_20}
	\vspace{10pt}
\end{figure*}

\begin{figure*}[!t]
	\centering
	\begin{tabular}{cc}
		\hspace{5.2mm}\includegraphics[trim = {5cm, 10cm, 5cm, 10cm}, width=.415\linewidth]{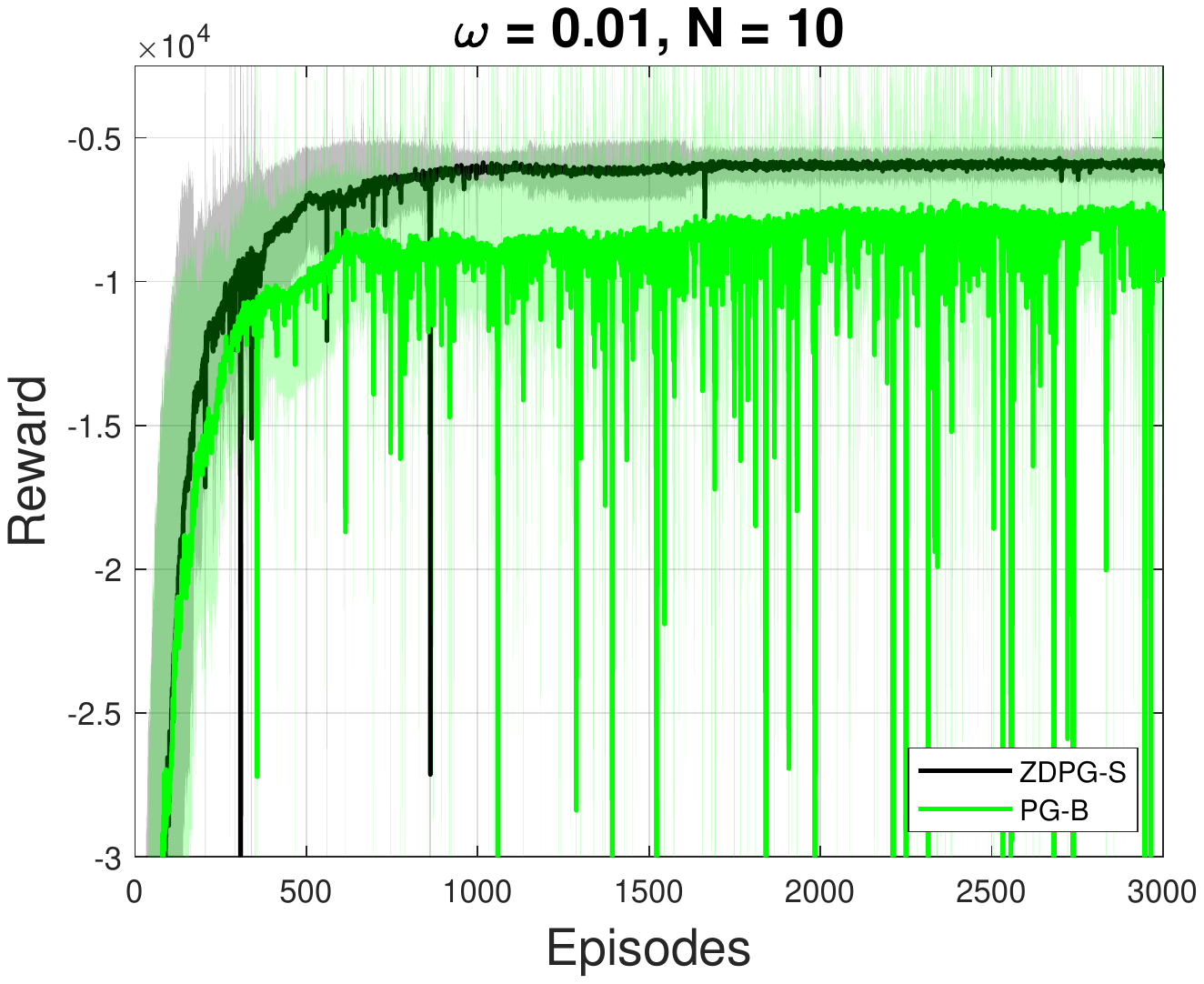}   &\hspace{8mm}
		\includegraphics[trim = {5cm, 10cm, 5cm, 10cm},width=.415\linewidth]{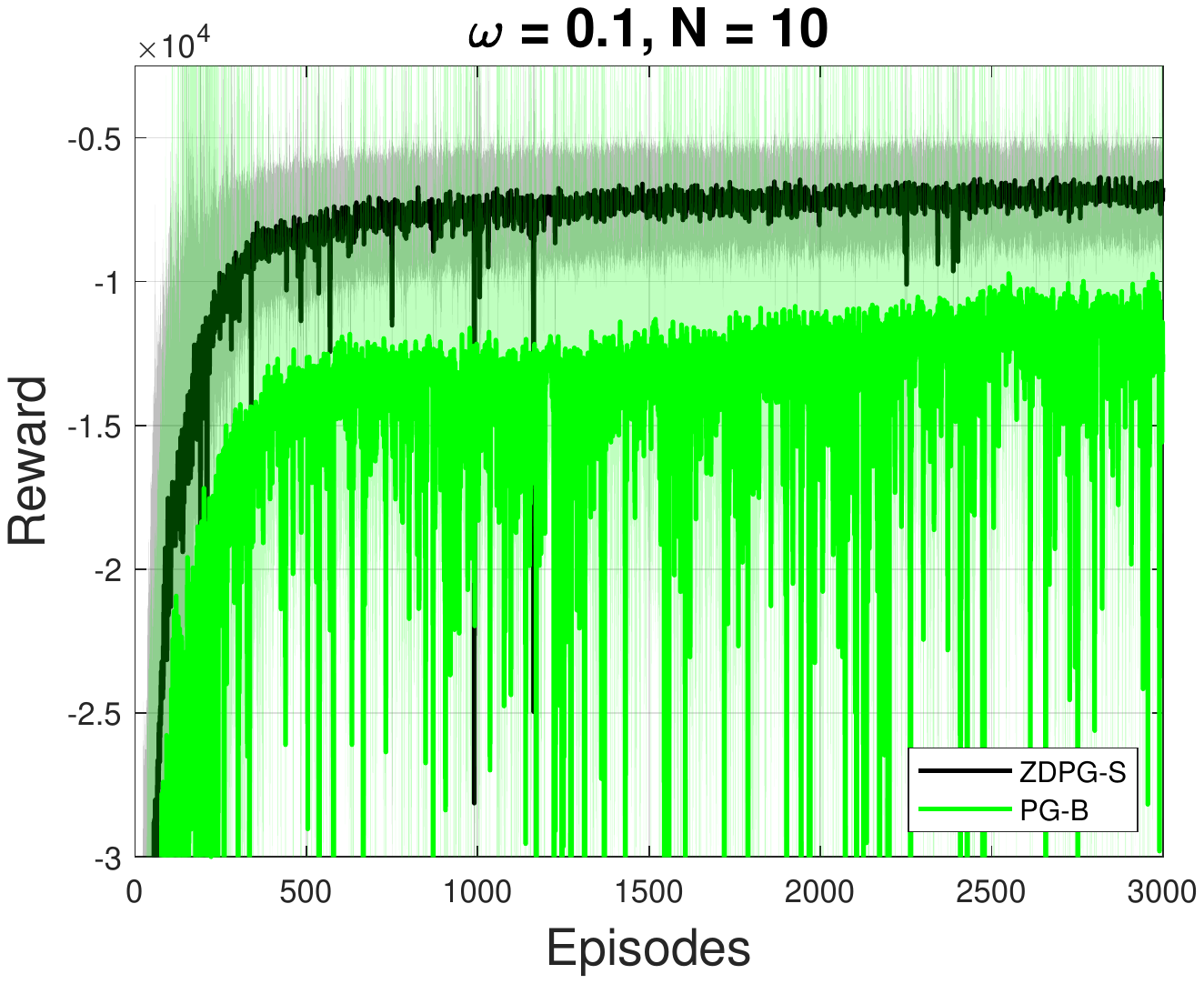} \\ \vspace{.5cm}
		~&~\\
	\end{tabular}
	\caption{Average reward per episode of length $T = 100$ with confidence bounds over 50 trials.}\vspace{2mm}
	\label{fig:LC_100}
	\vspace{-10pt}
\end{figure*}

\begin{figure*}[!t]
	\centering
	\begin{tabular}{ccc}
		\hspace{3mm}\includegraphics[trim = {5cm, 10cm, 5cm, 10cm}, width=.274\linewidth]{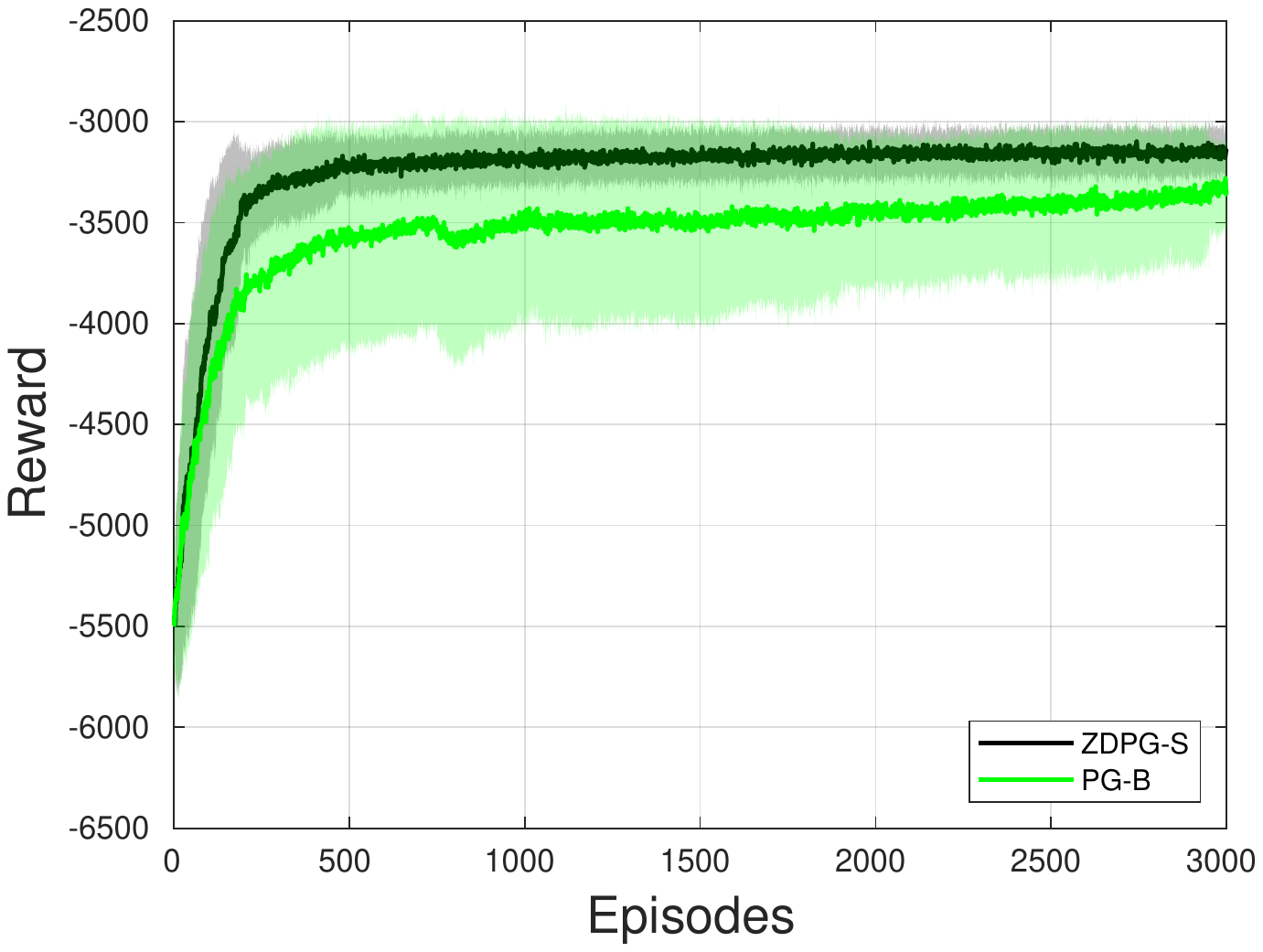}   &\hspace{3.5mm}
		\includegraphics[trim = {5cm, 10cm, 5cm, 10cm},width=.274\linewidth]{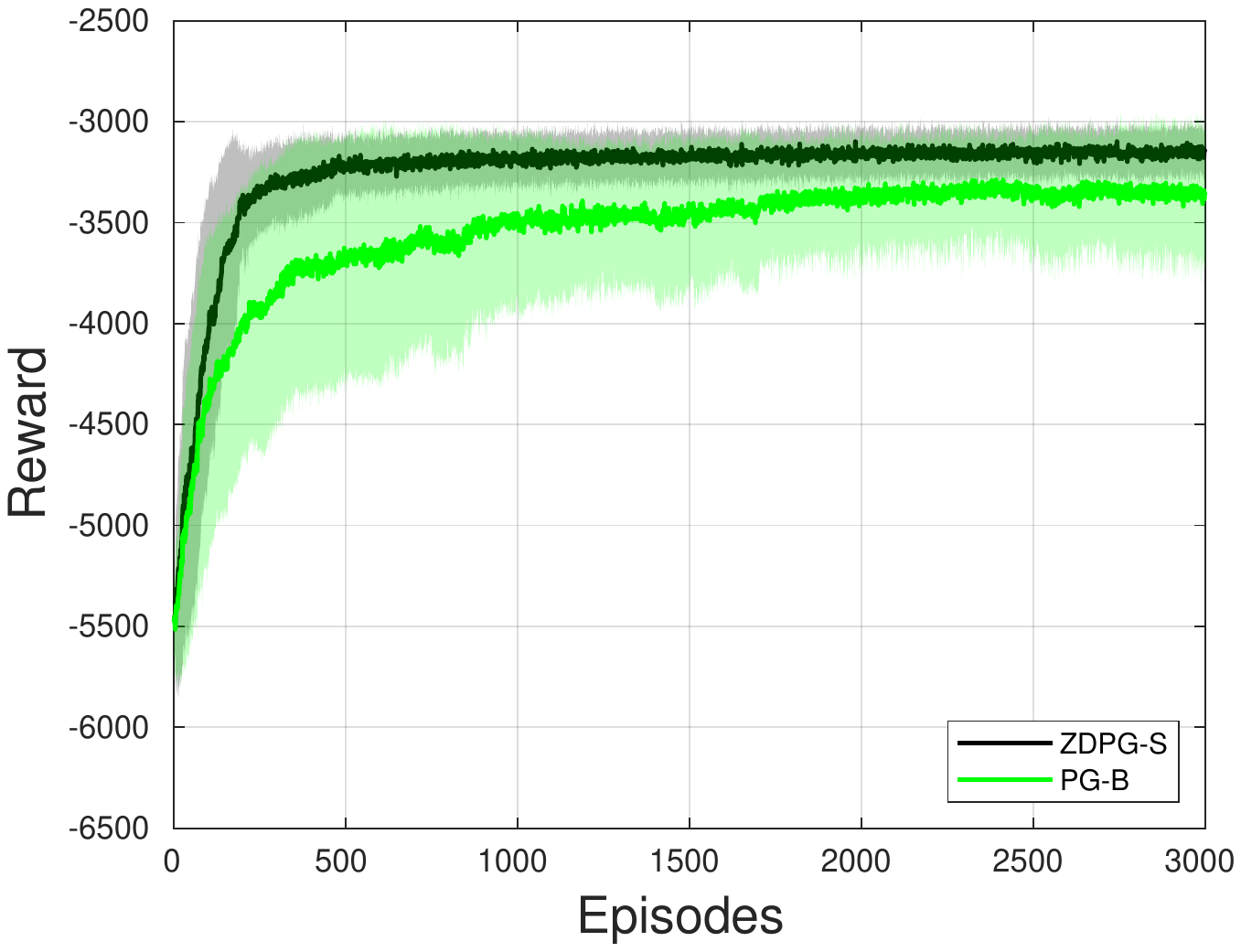} &\hspace{3.5mm}
		\includegraphics[trim = {5cm, 10cm, 5cm, 10cm},width=.274\linewidth]{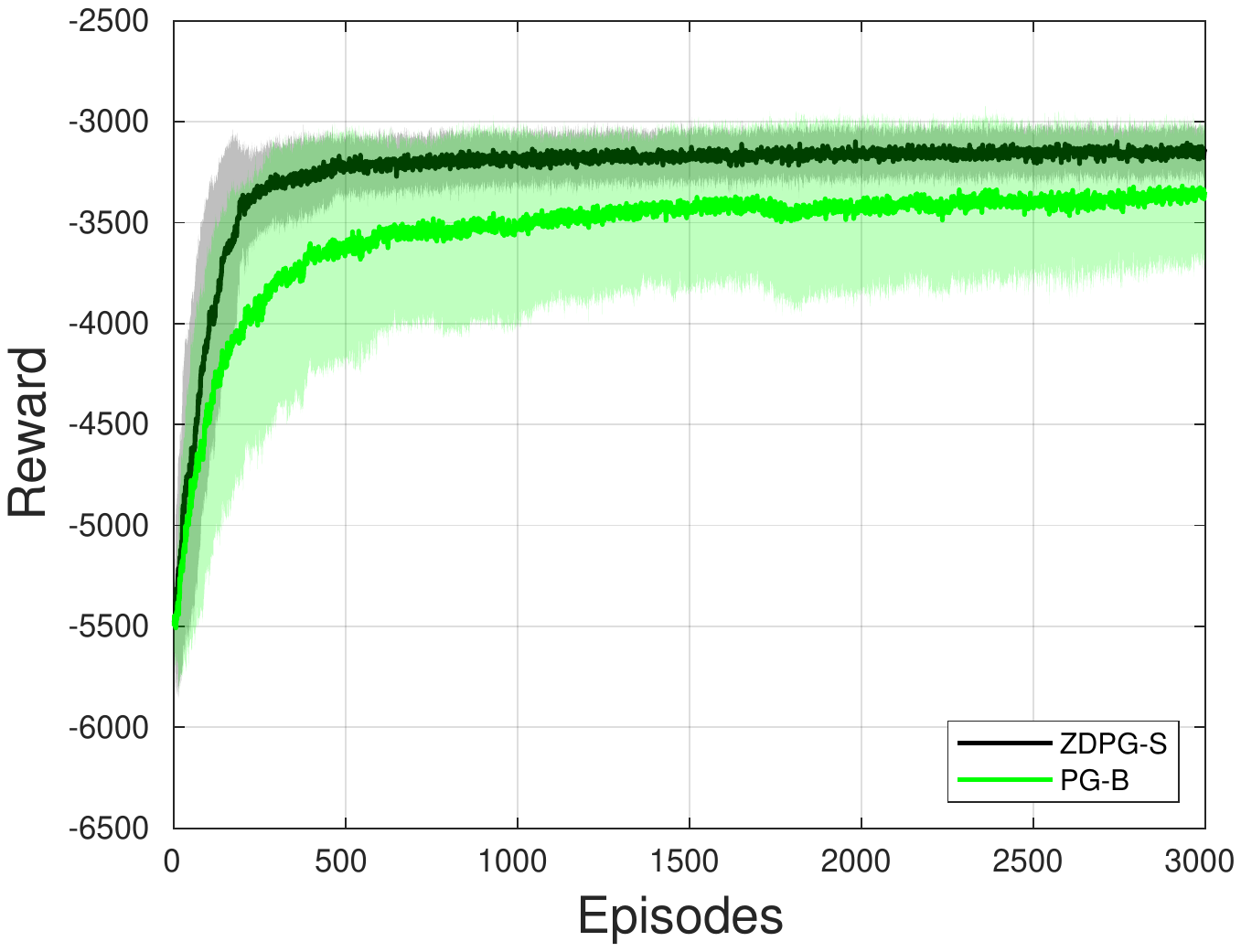} \\ \vspace{.4cm}
		~&~&~\\
		\small (a) & \small (b) & \small (c)
	\end{tabular}
	\caption{Average reward per episode of length $T = 20$ with confidence bounds over 50 trials. System noise $\omega = 0.01$ with no MC variance reduction ($N = 1$). Variance reduction is applied \textit{only} on value-function estimates to approximate the true (deterministic) baseline exclusively for PG-B, with (a) $20$, (b) $200$, and (c) $600$ rollouts.}
	\label{fig:true_baseline}
\end{figure*}
In section \ref{sec:Numericals}, we presented the learning curves for our algorithm implemented on a navigation problem with and without system noise. We showed that either with or without system noise, ZDPG and ZDPG-S enjoy a substantial advantage over PG with baseline (PG-B), and of course standard PG, especially without MC variance reduction, which is of importance in various practical settings.
The main purpose of this section is to confirm that the advantage of ZDPG does not disappear as the noise in the system further increases; in fact, it becomes even more drastic. We show this by comparing the learning curves on our navigation example for system noise variance $\omega = 0.01$ and $\omega = 0.1$, for different values of Monte-Carlo variance reduction $(N)$, and policy evaluation rollout length $(T)$. In this set of simulations, we only compare ZDPG-S with PG-B, since these methods seem to be empirically superior to ZDPG and PG, respectively.  

In Figure \ref{fig:LC_20}, the evaluation rollout length is set to $T = 20$. The smoothing parameter for ZDPG-S is $\mu  = 0.5$ for the $\omega = 0.01$ system noise case and $\mu = 0.6$ for the $\omega = 0.1$ system noise case. Analogously, for PG-B, we select action variance as $0.025$ for both levels of system noise as this value showed the most consistent performance throughout this line of experiments (by trial-and-error). For all simulations and for both methods, the learning rate was set to $\alpha = 10^{-7}$. Figure \ref{fig:LC_20} (top) shows that, indeed, ZDPG-S remains advantageous over and consistently outperforms PG-B when the system noise is increased. More interestingly, ZDPG-S shows significant advantage over PG-B in terms of noise resilience especially when there is no MC variance reduction at all (i.e. $N = 1$), which, as also stated above, is a case of interest in most practical scenarios. 

Figure \ref{fig:LC_100} shows the learning curves for both methods when the evaluation length is set to $T = 100$. While the performance of both ZDPG-S and PG-B is more sensitive over such a longer evaluation horizon, the plots show that ZDPG-S experiences much more stability  in solving the navigation problem for both levels of system noise. In contrast, PG-B becomes significantly unstable, resulting in trajectories that either do not necessarily reach the target (see Figure \ref{fig:traj}), or reach the target after a phase of prolonged, suboptimal wandering.

Because PG-B relies on the estimate ($\hat Q(s,a) - \tilde{V}(s)$),
%
%
MC variance reduction (whenever $N>1$) is essentially applied to both parts of the involved two-point difference for the results presented in Figure \ref{fig:LC_20} (and analogously for the two-point gradient estimates in ZDPG). In that way, the comparison between ZDPG and PG-B is fair for every $N\ge1$. Still, it would be interesting to compare ZDPG-S with PG-B, when in the latter 
a more accurate version of the corresponding true value function is employed, resembling a standard (non-stochastic) PG with baseline. To do this, we conducted the same experiment as in Figure \ref{fig:LC_20} for the low system noise case ($\omega = 0.01$) and with no MC reduction ($N=1$), but where only the baseline $\tilde{V}(s)$ is constructed as a \textit{multiple-rollout} estimate of the associated value function at state $s$ (i.e., an average).
We note, though, that this procedure results in a PG-B algorithm which is much more wasteful in terms of resources, whereas at the same time its comparison with ZDPG is extremely unfair, in favor of ZDPG.
This resulting plots are shown in Figure \ref{fig:true_baseline}. It very interesting to see that, even though the performance of PG-B slightly improves with the consideration of multiple-rollouts for the corresponding value function estimate, it still performs rather poorly compared with ZDPG-S, which of course demands just two system rollouts per episode.

Again, we would like to reiterate that the success of ZDPG comes from the non-heuristic use of deterministic policies, in contrast to random policies essential in log-trick-based standard (baseline) PG methods. At least with regards to our simulation setting, removing this level of randomness from the evolution of the system during training allows the agent to reach a solution quickly and consistently.